%% file: main.tex
\documentclass[5p,twocolumns]{elsarticle}

\usepackage{amssymb}
\usepackage{amsmath}
\usepackage{amsfonts}
\usepackage{amsthm}

\usepackage[T1]{fontenc}
\usepackage[polish,english]{babel}
\usepackage[utf8]{inputenc}

\usepackage{caption,subcaption}
\usepackage{microtype}
\usepackage{graphicx}
\usepackage{booktabs} 
\usepackage[linesnumbered,ruled,vlined]{algorithm2e}
\SetKwComment{Comment}{$\triangleright$\ }{}
\SetCommentSty{mycommfont}

\usepackage{dsfont}

\usepackage{cuted}
\usepackage{xcolor,colortbl}

\usepackage{hyperref}
\usepackage{tabularx}

\usepackage{float}
\usepackage{multirow}
\usepackage{dirtytalk}
\usepackage[toc,page]{appendix}

\newcommand{\FrameworkName}{FILDNE}
\newcommand{\MethodNameBasic}{FILDNE}
\newcommand{\MethodNameGeneralized}{k-FILDNE}
\newcommand{\MethodNameBasicLC}{fildne}
\newcommand{\MethodNameGeneralizedLC}{k-fildne}
\newcommand{\FrameworkFullName}{Framework for Incremental Learning of Dynamic Networks Embeddings}

\DeclareMathOperator*{\argmin}{argmin}

\usepackage{color,soul}
\usepackage{ulem} 

\theoremstyle{remark}
\newtheorem{defn}{Definition}

\journal{Knowledge Based Systems}

\begin{document}

\begin{frontmatter}

\title{FILDNE: A Framework for Incremental Learning of Dynamic Networks Embeddings}

\author[Pwr]{Piotr Bielak}
\author[Pwr]{Kamil Tagowski}
\author[Pwr]{Maciej Falkiewicz}
\author[Pwr]{Tomasz Kajdanowicz}
\author[Pwr,UND]{Nitesh V. Chawla}

\address[Pwr]{Department of Computational Intelligence, Wroclaw University of Science and Technology, Poland}
\address[UND]{Department of Computer Science and Engineering, University of Notre Dame, Notre Dame, IN, USA}

\address{}

\begin{abstract}
Representation learning on graphs has emerged as a powerful mechanism to automate feature vector generation for downstream machine learning tasks. The advances in representation on graphs have centered on both homogeneous and heterogeneous graphs, where the latter presenting the challenges associated with multi-typed nodes and/or edges. In this paper, we consider the additional challenge of evolving graphs. We ask the question whether the advances in representation learning for static graphs can be leveraged for dynamic graphs and how? It is important to be able to incorporate those advances to maximize the utility and generalization of methods. To that end, we propose the Framework for Incremental Learning of Dynamic Networks Embedding (FILDNE), which can utilize any existing static representation learning method for learning node embeddings, while keeping the computational costs low. FILDNE integrates the feature vectors computed using the standard methods over different timesteps into a single representation by developing a convex combination function and alignment mechanism. 
Experimental results on several downstream tasks, over seven real world data sets, show that FILDNE is able to reduce memory and computational time costs while providing competitive quality measure gains with respect to the contemporary methods for representation learning on dynamic graphs. 
\end{abstract}

\begin{keyword}
Representation learning \sep Dynamic graph embedding \sep Incremental network embedding
\end{keyword}

\end{frontmatter}

\section{Introduction}

Learning embeddings from networks or graphs is pervasive with a sundry of applications across various fields, including social networks \cite{tang2015line, grover2016node2vec, huang2019network, xu2020manifold}, biological networks \cite{nelson2019embed, yue2020graph}, molecular networks \cite{you2018graph, zang2020moflow}, spatial networks \cite{xu2020ge, wu2019graph}, citation networks \cite{tang2015line, dong2017metapath2vec, gao2020community}, transportation networks \cite{geng2019spatiotemporal} and many others. These embeddings are generally learned in an unsupervised fashion, providing an automated way of discovering dynamic features and enabling several downstream inductive learning tasks (and in some cases transductive learning tasks as well).

The vast majority of graph embedding methods are devoted to so-called static networks, whose structure does not evolve over time. However, in most real-world scenarios, one has to deal with changes in the data, e.g., updates of node attributes and structural adjustments, like addition or removal of edges (links) and nodes. On the one hand, dynamics can infrequently occur, which can be easily addressed with static approaches. On the other hand, there are streams of temporal events that constitute constantly evolving networks. The latter case requires dedicated, computationally, and memory-wise efficient solutions.  Thus, a key challenge remains: how to effectively learn network embeddings on dynamic networks? While recent works have addressed incremental learning of embeddings on dynamic graphs, in this paper, we consider a framework that encompasses existing methods for learning embeddings and enables them for a dynamic environment. 

\begin{figure}[ht]
    \centering
    \includegraphics[width=\columnwidth]{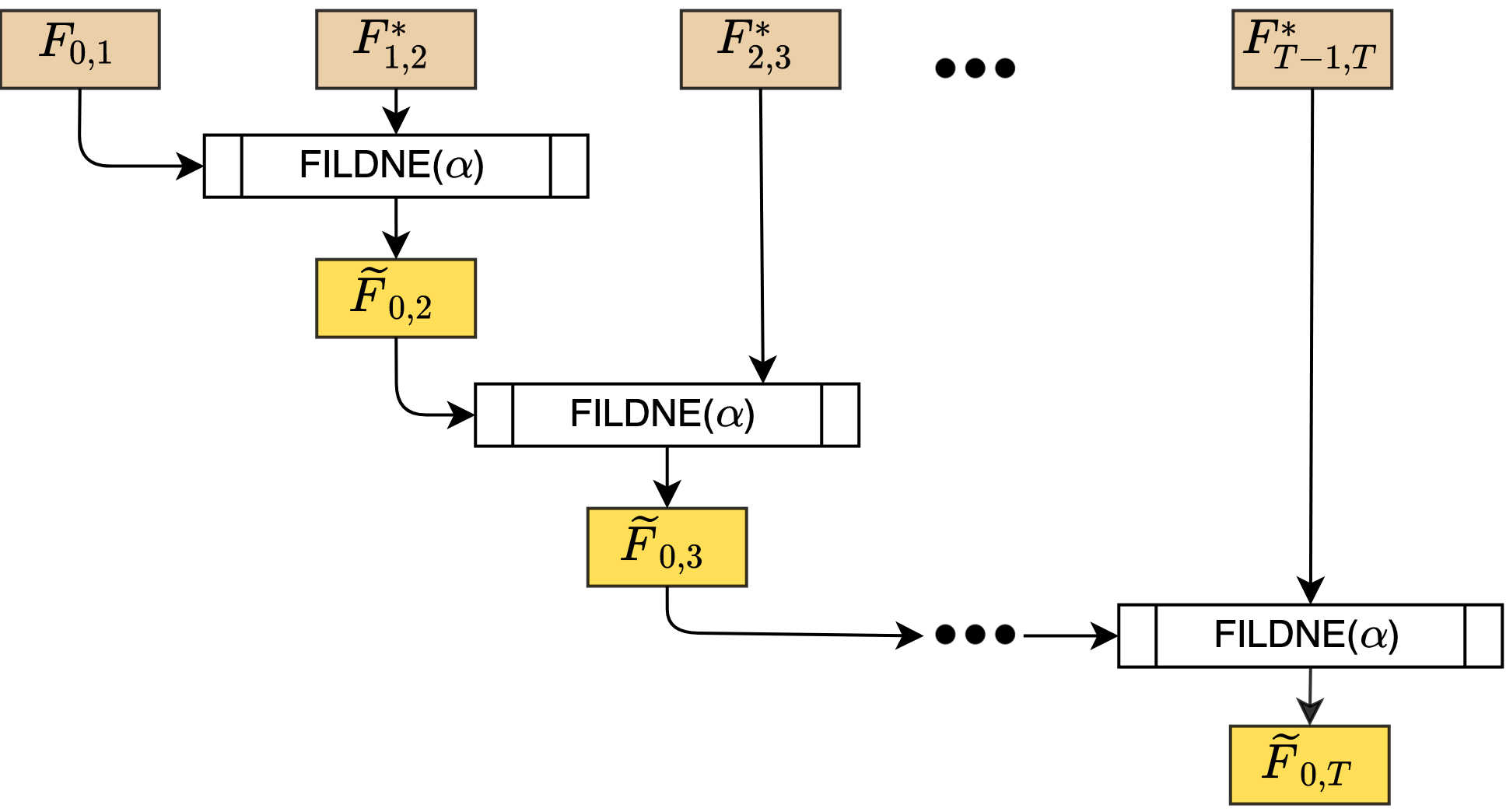}
    \caption{The {\MethodNameBasic} method applied on graph stream. At the beginning, {\FrameworkName} composes of two embeddings computed by the Base embedding method. Next, with each new snapshot,  {\FrameworkName} composes the aligned version of the current embedding $F_{t-1,t}^*$ with the output of the previous iteration $\tilde{F}_{0,t}$. }
    \label{fig:basic-icmen-inference}
\end{figure}

Let us consider two conceptually different aspects of \emph{Dynamic Graph Embedding}. The first is a naive one, where each new data batch triggers the computation of a new representation for historical data, completely disregarding previous feature vectors. The second one is an incremental learning paradigm, where both the time and storage costs are reduced by updating embeddings based on new temporal events. Incremental learning algorithms might specify a new objective function and could also be constrained to the types of problems that they could be applied to \cite{chami2020machine}. However, there is also a need for a framework that is able to incorporate the existing works for embeddings and implement them in an incremental manner, allowing for time and space costs to be reduced while retaining the quality of the base method.

\begin{figure*}[ht]
    \centering
    \begin{subfigure}[t]{0.45\textwidth}
         \centering
         \includegraphics[width=\textwidth]{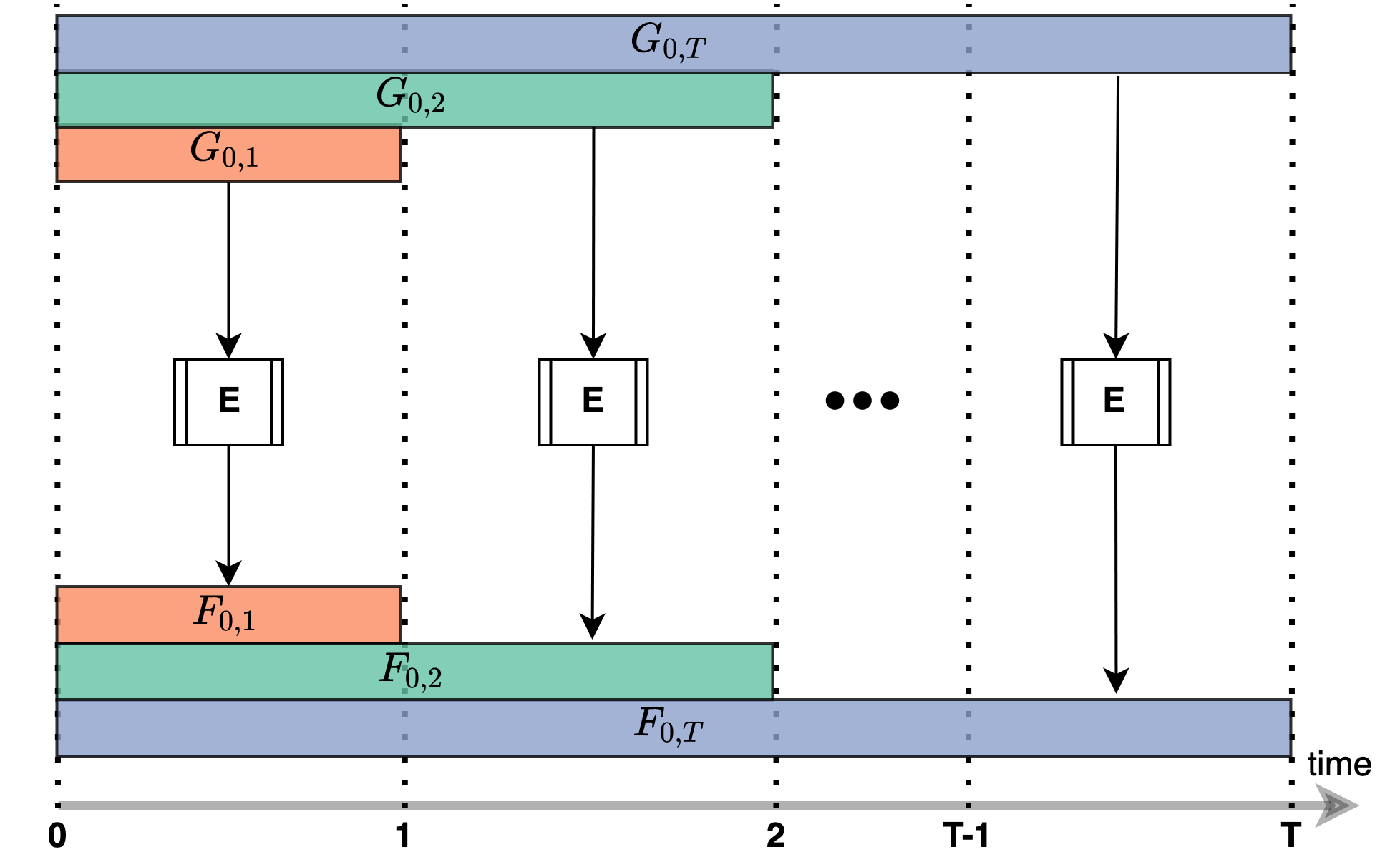}
         \caption{Using a static or temporal embedding method (\textbf{E}) requires to fully retrain the method on every new batch using cumulative snapshots.}
     \end{subfigure}
     \hfill
    \begin{subfigure}[t]{0.45\textwidth}
         \centering
         \includegraphics[width=\textwidth]{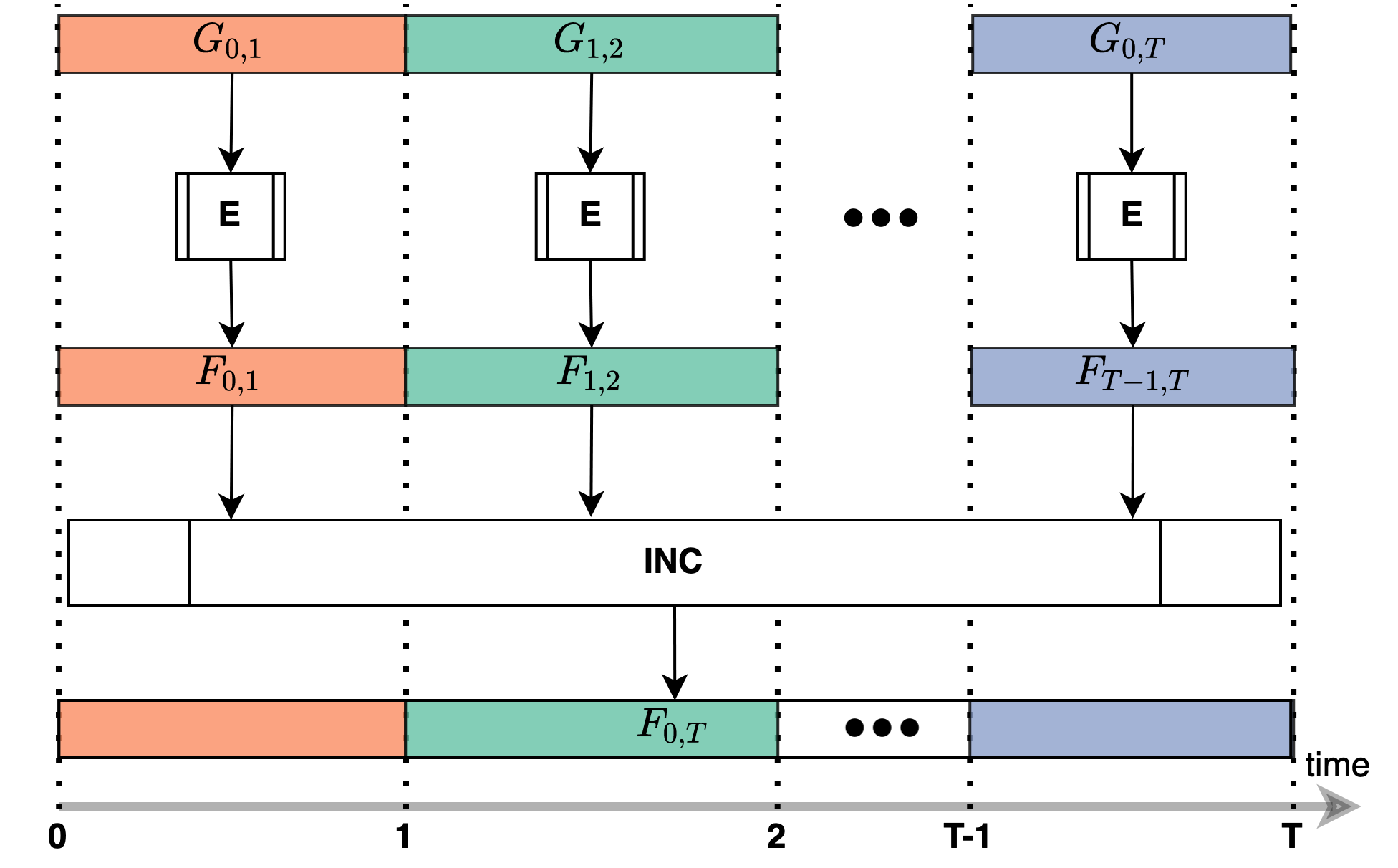}
         \caption{Using an incremental embedding method (\textbf{INC}) it is possible to reuse the already computed embeddings of non-overlapping snapshots and combine those into an embedding which describes the full event stream history.}
     \end{subfigure}
    
    \caption{Comparison of applications of traditional static network embedding methods and incremental methods in dynamic network embedding task.}
    \label{fig:traditional-and-icmen-embeddings}
\end{figure*}

\paragraph{This work} In this paper, we propose {\FrameworkName}, for incremental learning of embeddings of dynamic network. Our contribution can be summarized as follows:

\begin{enumerate}
\item We provide a method that utilizes historical embeddings and incrementally enhances them based on batched event stream (\emph{non-overlapping snapshots} for keeping the embeddings current; see Figure \ref{fig:traditional-and-icmen-embeddings}). We consider two variants: (a) the first one, {\MethodNameBasic},  recursively combines a pair of embeddings at each step using a hyper-parameter to steer the importance weighting (see Figure \ref{fig:basic-icmen-inference}; (b) the second one, {\MethodNameGeneralized},  combines a vector of $k$ embeddings at once using an \emph{automatic estimation method for importance weighting parameters}.

\item The proposed method can work with \emph{any} graph embedding method, including static and temporal graph embedding methods. Using a novel \emph{reference nodes selection scheme}, our method performs an  \emph{embedding alignment} step that allows us to apply convex combination despite rotations and translations of embedding spaces. Moreover, our framework is designed to work in an \emph{unsupervised manner} that does not require obtaining any additional class labels.

\item Through comprehensive empirical analyses that include link prediction, edge classification, and graph reconstruction tasks, we demonstrate that {\FrameworkName} allows \emph{reducing memory and computational time costs} while being competitive when compared to other streaming methods in terms of embedding quality. Our hyper-parameter sensitivity study (see Figure \ref{fig:basic-icmen-sensitivity}) shows how the balance of importance of past and recent events (data batches) influences the performance of the representation.
\end{enumerate}

The paper is organized as follows: in Section \ref{sec:related-work}, we present an overview of the related work in the domain, whereas in Section \ref{sec:notations}, we introduce several formal definitions and notation. Next, in Section \ref{sec:method}, we propose our {\FrameworkFullName} and provide its detailed description. Further, in Section \ref{sec:experiments} we report our extensive experiments results. Section \ref{sec:conclusion} concludes our work and outlines the directions for future work.

\section{Related work}
\label{sec:related-work}

Network embedding problem has attracted a lot of attention from the research community worldwide in recent years. Plenty of methods have been developed, each focused on a different aspect of network embedding, such as proximity, structure, attributes, learning paradigm, scalability, to name only a few \cite{Cai2018, Cui2019, chami2020machine}. In this section, we discuss several methods that are relevant to the scope of our paper. We summarize these in Table \ref{tab:methods}, where we adapt the taxonomy introduced in \cite{Cai2018}.

\subsection{Static network embedding}

The topic is covered in various embedding method families, among which we first discuss \emph{matrix factorization} based methods. \textbf{Locally Linear Embedding (LLE)} \cite{roweis2000nonlinear} and \textbf{Laplacian Eigenmaps (LE)} \cite{belkin2003laplacian} both aim to map a high-dimensional data point space to low-dimensional one based on the neighbourhood of points (first-order proximity). LLE reconstructs a linear weight matrix, and in LE eigenvectors over graph Laplacian are computed. \textbf{Large-scale Information Network Embedding (LINE)} \cite{tang2015line} extended these approaches by additionally preserving second-order proximities in the graph. Nodes with similar neighborhoods end up lying closer in the embedding space. \textbf{High Order Proximity preserved Embedding (HOPE) }\cite{Ou2016} aims to sustain asymmetric proximities of nodes in the graph, in contrast to LINE, where proximities were symmetric.

The next group is methods which are based on \emph{random-walks}. In \textbf{DeepWalk} \cite{perozzi2014deepwalk} random-walks sampled over the graph are fed to the skip-gram model adapted from Word2Vec \cite{Mikolov2013EfficientEO}. \textbf{Node2Vec} \cite{grover2016node2vec} was an improvement over DeepWalk, where authors introduced parameters $p$ and $q$ control both depth-first search and breadth-first search like behavior.

Another approach utilizes \emph{Graph Convolutional Networks} architecture -- \textbf{Deep Graph Infomax (DGI)} \cite{velickovic2019deep} is an unsupervised method, which relies on maximizing mutual information between patch representations (obtained by Graph Convolutional Network-based encoder layer), and corresponding high-level summaries of graphs (obtained by readout function). All of the approaches mentioned above are capable of processing static networks only and are not open. 

\subsection{Temporal and Dynamic Network Embedding}

In Temporal Network Embedding methods, we aim to preserve the temporal properties of the network. On the other hand, Dynamic Network Embedding focuses on providing up-to-date embedding for evolving graphs. We can distinguish \emph{online} approaches that update embedding with every new edge arrival and \emph{incremental} approaches that process events in batches. The majority of the methods satisfy both of these objectives -- temporal and dynamic. Here we give a brief overview of the most prominent ones. A popular trend found in the literature is to build upon the framework of random walks with the skip-gram model. \textbf{Continuous-Time Dynamic Network Embedding (CTDNE)} \cite{Nguyen2018} introduced temporal walks that traverse edges according to their timestamps instead of performing random-walks statically. The original version of the method focused on the temporal aspect of embedding, while in the follow-up work, the authors introduce an online version of the algorithm \cite{lee2019dynamic} that produces a new portion of temporal random walk for upcoming events and then updates the model. This direction is further extended in \textbf{Dynnode2vec} \cite{mahdavi2018dynnode2vec} architecture, where for each timestamp, a random-walk is sampled, only for nodes marked as evolving in terms of new edges. \textbf{Global Topology Preserving Dynamic Network Embedding (GloDyNE)} \cite{hou2020glodyne} also follows a similar schema with incrementally updating skip-gram model, but they differ in the method of selecting nodes to perform new random-walks. They partition the graph into $k$ sub-networks for each timestamp, and for each sub-network, they randomly select one node based on calculated probability distribution within the sub-network. After obtaining the representative nodes list, they perform new random-walks over a new snapshot for representative nodes and update the skip-gram model. An interesting approach was introduced in \textbf{Online-Node2Vec} models \cite{Beres2019}: StreamWalk and SecondOrder. StreamWalk follows a temporal random walk procedure like CTDNE but differs in the edge sampling scheme. In their second model SecondOrder, they use MinHash fingerprinting to approximate Jaccard node similarity instead of performing temporal random-walks that reduce the method's complexity. In \textbf{Weg2vec} \cite{torricelli2020weg2vec}, instead of embedding nodes, representations of events are learned by introducing a weighted neighborhood edge sampling strategy. Finally in \textbf{tNodeEmbed} \cite{singer2019node} for each timestamp a new embedding is calculated using the Node2vec method. Embeddings are aligned between timestamps before they are fed as an input to the Long Short Term Memory (LSTM) model, which performs an end-to-end task.

In contrary to random-walk-based methods \textbf{Dyngraph2Vec} \cite{goyal2020dyngraph2vec} extends Autoencoder (AE) architecture to capture to the evolving structure of temporal networks providing a purely neural-network-based approach. They present three variants of the model: dyngraphAE, dyngraphRNN, and dyngraphAERNN, which differ in how they represent input neighbor vector -- dyngraphAE uses fully connected layers, dyngraphRNN uses LSTM layers, and DyngraphAERNN uses fully connected layers followed by LSTM layers. 

The authors of \cite{trivedi2018structural} propose a framework (further referred to as \textbf{LCF}) that is based on the linear combination of embeddings from consecutive snapshots. Building upon a similar paradigm, we provide an insightful framework with differences discussed in Section \ref{sec:methods_comparison}.

\subsection{Network embedding alignment} 
Network embedding alignment problem arises when combining embeddings from subsequent runs or comparing representations from different graphs. \cite{grave2019unsupervised,chen2020consistent} identify cross-graph node similarities by jointly solving two optimization problems: they use the Sinkhorn algorithm to match node correspondence and find a linear transformation of one of the embeddings by solving Orthogonal Procrustes. \cite{singer2019node} uses Orthogonal Procrustes to see a transformation matrix between embeddings of two consecutive timestamps (as we do). They use all common nodes between timestamps to form the matrix (which differs from our method). \cite{liu2016aligning} uses alignment as an integral part of the model, improving the individual learning of embeddings. Embeddings are aligned at anchor nodes (that indicate the same users across two networks) and introduce soft-constraint for non-anchor nodes. Other approaches utilize generative adversarial networks \cite{derr2019deep, zhang2020wasserstein} to align embeddings. \cite{zhang2020wasserstein} solution is based on Wasserstein GAN to produce cross-lingual embedding mapping. \cite{derr2019deep} utilized GAN architecture in which they aim to obtain both sides' transformation using cycle consistency loss.

\begin{table*}[ht]
\caption{Graph node embedding methods comparison. Methods marked in bold are the ones evaluated in our experiments.}
\label{tab:methods}
    \begin{center}
        \begin{small}
            \begin{sc}
                \scalebox{0.9}{
                    \begin{tabular}{l|l|l|l|l|l}
                        \toprule
                         Method & Static& Temporal & Dynamic & Network & Taxonomy \\
                          &&&& Alligment &\\
                         \midrule
                         \textbf{LLE'00} \cite{roweis2000nonlinear}& $\surd$  & $\times$ & $\times$ & $\times$ & Matrix Factorization\\
                         \textbf{LE'03} \cite{belkin2003laplacian} & $\surd$  & $\times$ & $\times$ &  $\times$ & Matrix Factorization \\ 
                         \textbf{LINE'15} \cite{tang2015line} & $\surd$  & $\times$ & $\times$ & $\times$ & Edge Reconstruction \\
                         \textbf{HOPE'16} \cite{Ou2016} & $\surd$  & $\times$ & $\times$ & $\times$ & Matrix Factorization\\
                         \textbf{DGI'19} \cite{singer2019node} & $\surd$  & $\times$ & $\times$ & $\times$ & Graph Convolutional Networks\\
                         \textbf{DeepWalk'14} \cite{perozzi2014deepwalk}& $\surd$  & $\times$ & $\times$ & $\times$ & Random Walk, Skip-Gram\\
                         \textbf{Node2Vec'16} \cite{grover2016node2vec}& $\surd$  & $\times$ & $\times$ &  $\times$ & Random Walk, Skip-Gram \\
                         \midrule
                         \textbf{CTDNE'18, 19} \cite{Nguyen2018, lee2019dynamic} & $\times$ & $\surd$ & $\surd$ & $\times$ & Temporal Random Walk, Skip-Gram\\
                         \textbf{tNodeEmbed'19} \cite{singer2019node}& $\times$ & $\surd$ &  $\surd$ &  $\surd$ & Neural Networks, Random Walk, Skip-Gram\\
                         \textbf{StreamWalk'19} \cite{lee2019dynamic}& $\times$ & $\surd$ & $\surd$ & $\times$ & Temporal Random Walk \\
                         \textbf{SecondOrder'19} \cite{lee2019dynamic}& $\times$ & $\surd$ & $\surd$ & $\times$ & Temporal Random Walk\\
                         \textbf{dyngraph2vec'20} \cite{goyal2020dyngraph2vec} & $\times$ & $\surd$& $\surd$  & $\times$ & Neural Networks \\
                         dynnode2vec'18 \cite{mahdavi2018dynnode2vec}& $\times$ & $\surd$& $\surd$  & $\times$ & Random Walk, Skip-Gram\\
                         weg2vec'20 \cite{torricelli2020weg2vec}& $\times$ & $\surd$& $\times$  & $\times$ & Temporal Random Walk, Skip-Gram \\
                         GloDyNe'20 \cite{hou2020glodyne}& $\times$ & $\surd$& $\surd$  & $\times$ & Random Walk, Skip-Gram \\
                         LCF'20 \cite{trivedi2018structural}  & $\times$ & $\surd$& $\surd$ & $\surd$ & Dependent on base method \\
                         \midrule
                         \textbf{{\FrameworkName}'20 (Our)} & $\times$ & $\surd$& $\surd$ & $\surd$ & Dependent on base method \\
                         \bottomrule
                    \end{tabular}
                }
            \end{sc}
        \end{small}
    \end{center}
\end{table*}

\begin{figure}[H]
    \centering
    \includegraphics[width=\columnwidth]{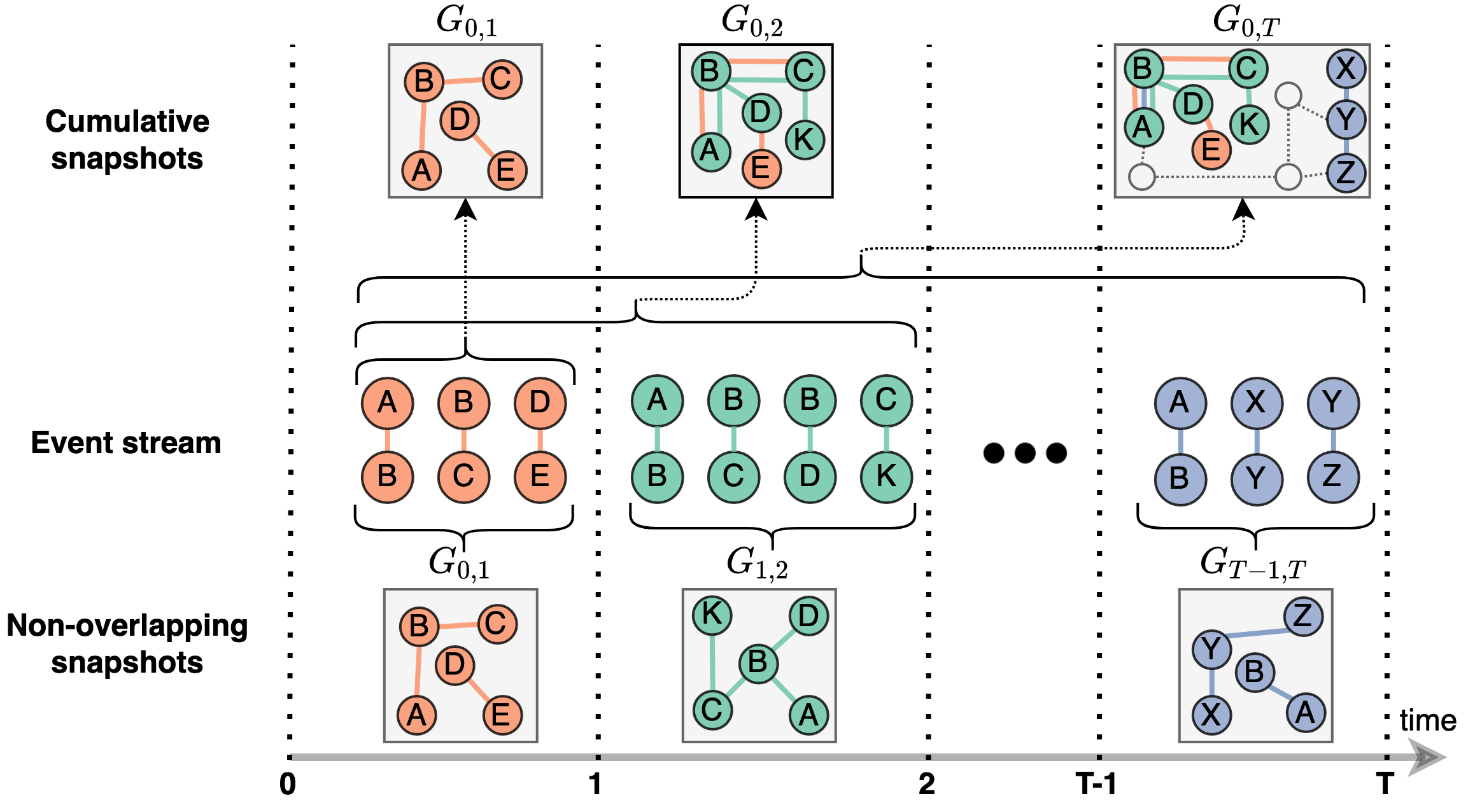}
    \caption{Event streams can be saved as a series of cumulative or non-overlapping graph snapshots. The first hold the full history from the very beginning, but at the cost of a relatively high memory footprint. Contrary, the latter ones are restricted to a given time interval, hence requiring less memory.}
    \label{fig:overlapping-nonoverlapping-snapshots}
\end{figure}

\section{Notation and problem definition}
\label{sec:notations}
The notation introduced in these and all the following is summarized in Table \ref{tab:notations}.

\begin{defn}[Static Network]
    A Network (Graph) is a pair $G = (V, E)$, where $V$ is a set of \emph{vertices} and $E = \{(u, v): (u, v) \in V \times V\}$, is a set of \emph{edges} connecting vertices. Both, the nodes and edges can posses assigned \emph{attributes}. A special kind of vertices' attributes are \emph{timestamps}, which lead to the next definition.
\end{defn}

\begin{defn}[Dynamic Network]
A Dynamic Network (Graph) is a triple $G = (V, E, ts)$ where $V$ and $E$ are sets of vertices and edges respectively and $ts: E \rightarrow \mathcal{R}$ is a function assigning \emph{timestamp} to each edge.

Working with such a network is inconvenient -- whenever we want to check the state of the graph at a given time $t$ we have to iterate over $E$. The solution would be to store it as a \emph{snapshot} $G_{0, t} = (V_{0, t}, E_{0, t}, ts)$, where $E_{0, t}$ is a set of edges with timestamps up to time $t$, while $V_{0, t}$ is the set of vertices associated with them. Further we arrange them in a sequence $[G_{0, 1}, G_{0, 2}, \ldots, G_{0, T}]$ of cumulative graphs, each associated with a \emph{time-index} in the range $[1; T]$, where $T$ denotes the maximal time-index. One might be interested in non-overlapping batches $[G_{0, 1}, G_{1, 2}, \ldots, G_{T-1, T}]$, where $G_{t, t+1}$ consists only of edges from $E_{t, t+1}$, created between $t$ and $t+1$.

For simplicity, we will mark snapshots with the end of interval whenever they cover a single time window, that is $G_{t-1, t} \equiv G_{t}$.

Dynamic Networks can be attributed in the same way as Static Networks are.
\end{defn}

\begin{defn}[Graph Stream]
A Dynamic Network is defined for a limited time interval $[0; T]$ specified by the youngest snapshot's time-index. A Graph Stream expands this definition for a potentially infinite stream of events (each connecting two nodes and represented as an edge) forming an infinite sequence of cumulative graphs $[G_{0, 1}, G_{0, 2}, \ldots]$ or equivalently non-overlapping ones $[G_{0, 1}, G_{1, 2}, \ldots]$ (see Figure \ref{fig:overlapping-nonoverlapping-snapshots}). The real-world applications of Graph Streams have to take resources limitation into account. Therefore the oldest history has to be forgotten or compressed.

A graph stream can be observed in an online (one edge at a time) or batched manner. In this paper, we would like to focus on the batched setting, remembering that the online setting can be interpreted as single-event batches.
\end{defn}

\begin{defn}[Static Network Embedding]
The aim is to find a mapping $f_G: V \rightarrow R^d$, $d \ll |V|$ such that the topological (proximity or structural) similarity of vertices in a static $G$ is preserved. The resulting embedding is marked as $F = f_G(V)$ and can be arranged as a ${|V| \times d}$ matrix, where each row denotes vector representation of a single node.
\end{defn}

\begin{defn}[Temporal Network Embedding]
The aim is to find a mapping $f_{G_{0, T}}: V_{0, T} \rightarrow R^d$, $d \ll |V|$ such that the temporal topological \cite{Marceau2001} (proximity or structural) similarity of vertices in a dynamic $G_{0, T}$ is preserved. The resulting embedding is marked as $F_{0, T} = f_{G_{0, T}}(V_{0, T})$ and can be arranged as a ${|V_{0, T}| \times d}$ matrix, where each row denotes vector representation of a single node.
\end{defn}

\begin{defn}[Dynamic Network Embedding]
As the network evolves one may be interested in evolving network embedding. We can distinguish two approaches -- naive and incremental one (see Figure \ref{fig:traditional-and-icmen-embeddings}). In the latter setting, we reuse previously computed embeddings $(F_{t-1}$, $f^{t-1}$, $\ldots$, $F_1$, $f^1)$ to obtain a representation $F_{0,t}$ updated with the most recent snapshot $G_t$, i.e. $F_{0, t} = f^{t}(G_t$, $G_{t-1}$, $F_{t-1}$, $f^{t-1}$, $\ldots$, $F_1$, $f^1)$. The motivation for incremental paradigm is to reduce computational cost of naive approach by updating nodes' embeddings. The training objective is preservation of the topological properties in $G_{0,t}$.
\end{defn}

\begin{defn}[Matrix alignment]
\label{def:matrix_alignment}
This can be seen as an instance of the \emph{orthogonal Procrustes} problem \cite{chen2020consistent}. Given matrices $A \in R^{n \times d}$ and $B \in R^{n \times d}$ with matching rows, we are interested in finding transformation matrix $Q$ that satisfies
\begin{equation}
    \argmin_{Q: Q^\intercal Q = I} ||B Q - A||_2^2.
\end{equation}
The solution is easily found as $Q^*=UW^\intercal$, where $U \Sigma W^\intercal$ is the Singular Value Decomposition (SVD) of $B^\intercal A$.
\end{defn}

\paragraph{Remark} For the completeness of the discussion on evolving networks, one should also consider such aspects as time attributes in the form of intervals, non-time attributes changing in time, and nodes appearing without an edge. However, these considerations go beyond the proposed method's scope, and we leave them as challenges for subsequent research.

\begin{table}[ht]
    \centering
    \caption{Symbols and notations}
    \label{tab:notations}
    \begin{tabularx}{\columnwidth}{cX}
        \toprule
        Symbol & Object \\
        \midrule
        $G$ & network / graph \\
        $G_{t}$ & snapshot of dynamic graph with nodes and edges occurring between $t-1$ and $t$ \\
        $G_{t_{1}, t_{2}}$ & snapshot of dynamic graph with nodes and edges occurring between $t_{1}$ and $t_{2}$ \\
        $V$ & set of nodes \\
        $V_{t_{1}, t_{2}}$ & set of nodes in dynamic graph $G_{t_{1}, t_{2}}$  \\
        $V_{t-1 \cap t}$ & set of common nodes between $G_{t-1}$ and $G_{t}$ \\
        $V_{ref}$ & set of reference nodes \\
        $E$ & set of edges \\
        $E_{t_{1}, t_{2}}$ & set of edges in dynamic graph $G_{t_{1}, t_{2}}$ \\
        $T$ & maximal time-index in Dynamic Network \\
        $F$ & embedding matrix \\
        $F(V')$ & embedding matrix of a subset of nodes $V' \subseteq V$ \\
        $F_{t}$ & embedding matrix for graph $G_{t-1, t}$ \\
        $F_{t_{1}, t_{2}}$ & embedding matrix for graph $G_{t_{1}, t_{2}}$ \\
        $F_{t_{1}, t_{2}}^{*}$ & aligned embedding for graph $G_{t_{1}, t_{2}}$ \\
        $\boldsymbol{F}_{k | t}^{*}$ & vector of $k$ aligned embeddings up to time $t$ \\
        $\tilde{F}_{0,t}$ & embedding matrix generated by {\FrameworkName}\\
        $\alpha$ & convex combination weight in {\MethodNameBasic}\\
        $\boldsymbol{\alpha}$ & vector of $k$ convex combinations weights in {\MethodNameGeneralized}\\
        $a(\cdot)$ & activity function \\
        $a_{t}^{(v)}$ & $v$'s activity in $G_{t}$ \\
        $s(\cdot, \cdot)$ & scoring function \\
        $S$ & scores of common nodes $V_{t-1 \cap t}$\\
        \bottomrule 
    \end{tabularx}
\end{table}

\section{Proposed {\FrameworkFullName}}
\label{sec:method}

The goal of {\FrameworkFullName} ({\FrameworkName}) is to find the embedding $F_{0,t}$ of the full graph $G_{0,t}$, without calculating it from all source data from period $[0, t]$, but based on already computed embeddings on historical data, i.e. $(F_{t-1}$, $\ldots$, $F_1)$ and the most recent snapshots $G_{t-1}$ and $G_t$. Let us note that our method does not require the mapping functions $(f^{t-1}$, $\ldots$, $f^1)$. {\FrameworkName} consists of 3 consecutive steps that are repeated with each new data portion arrival $G_{t-1,t}$. The methods come in two versions -- {\MethodNameBasic} and {\MethodNameGeneralized} --- therefore, step 3. has two variants. The difference between {\MethodNameBasic} and {\MethodNameGeneralized} method is that the former works in a pairwise manner $F_{0, t} = \text{\MethodNameBasic}(G_t$, $G_{t-1}$, $F_{t-1})$, while the latter operates on a vector of past embeddings $F_{0, t} = \text{\MethodNameGeneralized}(G_t$, $G_{t-1}$, $F_{t-1}$, $\ldots$, $F_{t-k})$, where $k$ is a parameter of the method.

\begin{figure*}
    \centering
    \begin{subfigure}[t]{0.32\textwidth}
         \centering
         \includegraphics[width=0.7\textwidth]{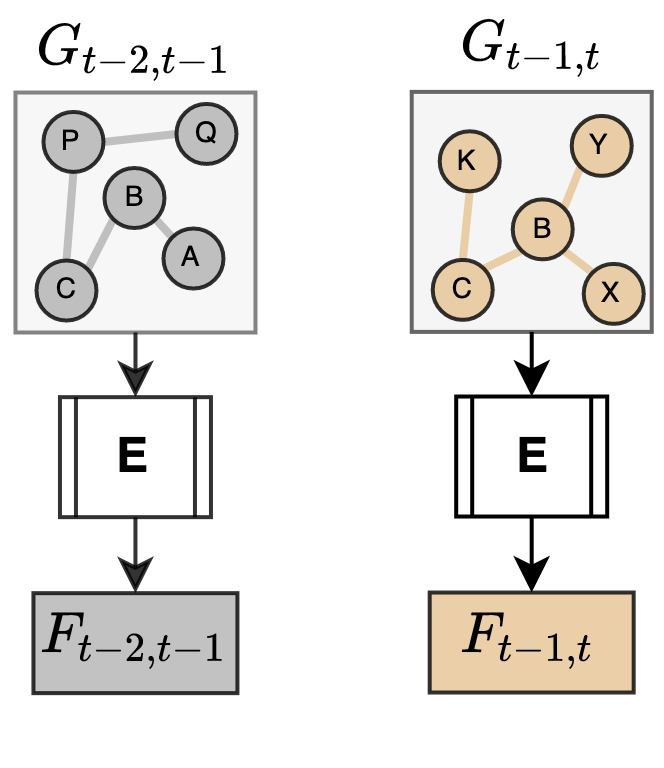}
         \caption{Batch embedding. Using Base method \textbf{E}, the first step is to compute the embedding $F_{t-1,t}$ for the most recent non-overlapping snapshot.}
         \label{fig:batch-embedding}
     \end{subfigure}
     \hfill
    \begin{subfigure}[t]{0.32\textwidth}
         \centering
         \includegraphics[width=0.7\textwidth]{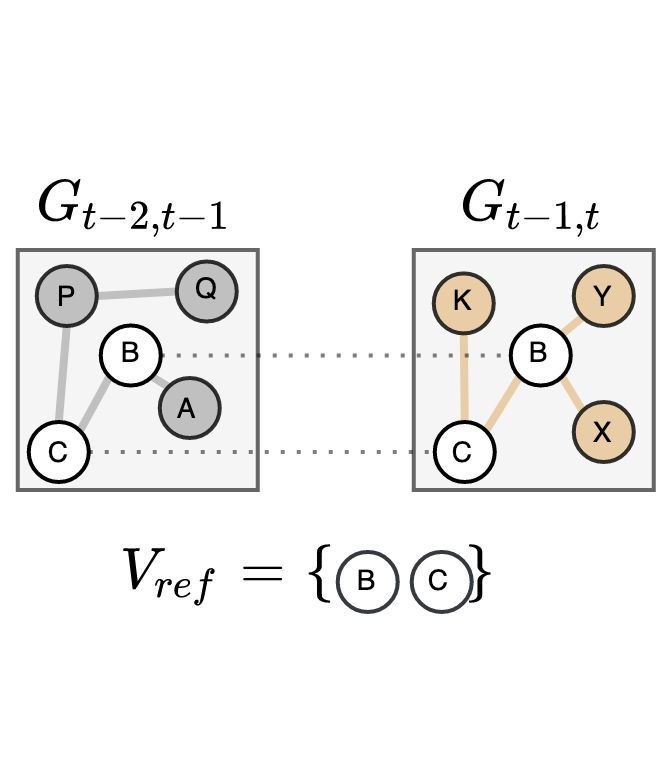}
         \caption{Reference node selection. Based on node activities $a(\cdot)$ from subsequent snapshots, a scoring function $s(\cdot, \cdot)$ is applied. A given node selection scheme uses those node scores, to obtain the reference nodes set $V_{ref}$.}
         \label{fig:reference-nodes-selection}
     \end{subfigure}
     \hfill
     \begin{subfigure}[t]{0.32\textwidth}
         \centering
         \includegraphics[width=0.7\textwidth]{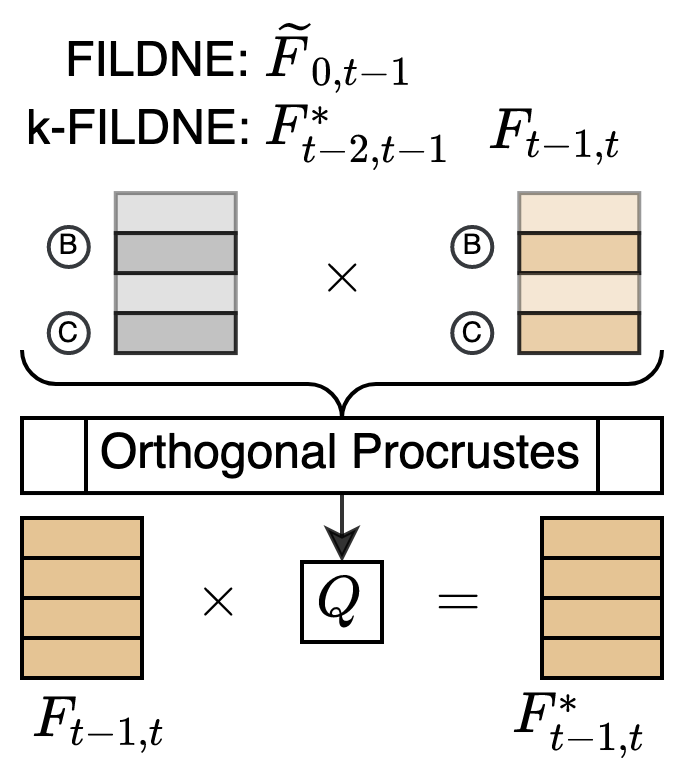}
         \caption{Embedding alignment. Solving the Orthogonal Procrustes problem for embeddings of reference nodes allows to obtain a transformation matrix $Q$, which is multiplied by $F_{t-1, t}$. {\MethodNameBasic} uses matrix $F_{0, t-1}(V_{ref})$ and {\MethodNameGeneralized} $F_{0, t-1}(V_{ref})$}
         \label{fig:embedding-alignment}
     \end{subfigure}
     
    \begin{subfigure}[t]{0.4\textwidth}
        \centering
        \includegraphics[width=\columnwidth]{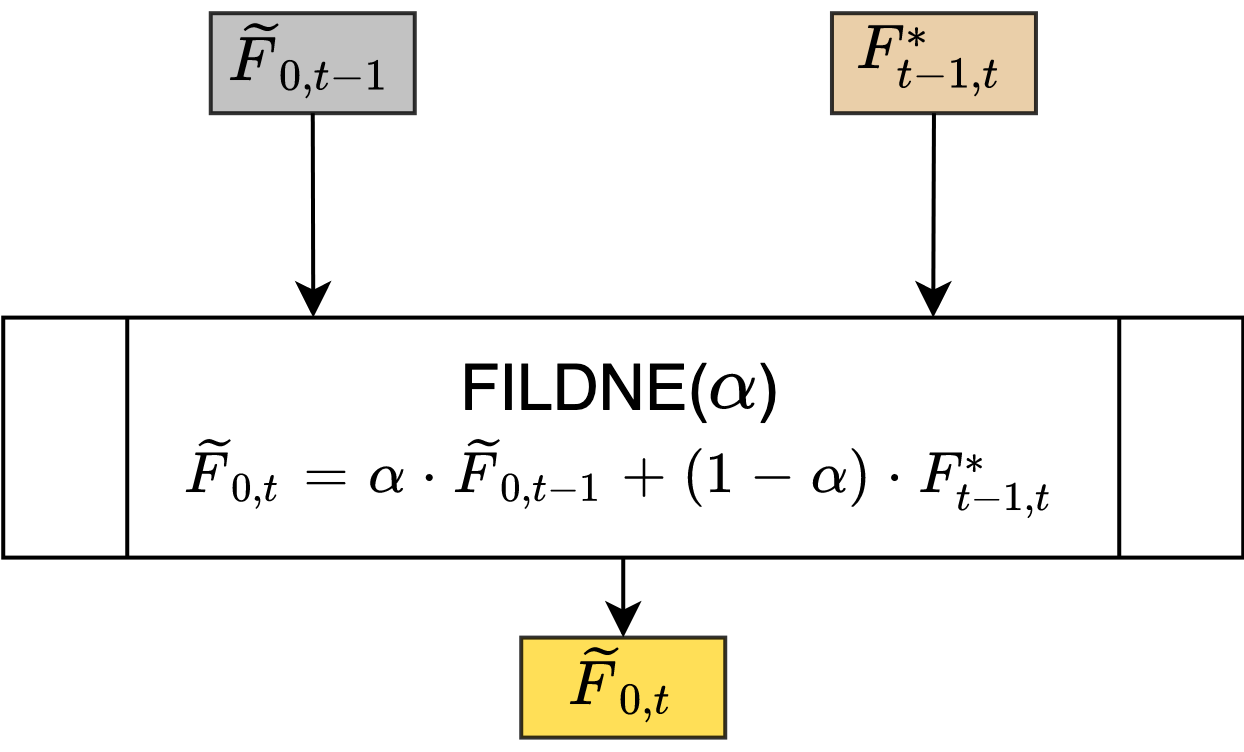}
        \caption{Single composition step of {\MethodNameBasic}. The previously composed embedding matrix $\tilde{F}_{0,t-1}$ is combined with the aligned embedding matrix $F^*_{t-1, t}$ using parameter $\alpha$.}
        \label{fig:bi-composition}
    \end{subfigure}
    \hfill
    \begin{subfigure}[t]{0.4\textwidth}
        \centering
        \includegraphics[width=\columnwidth]{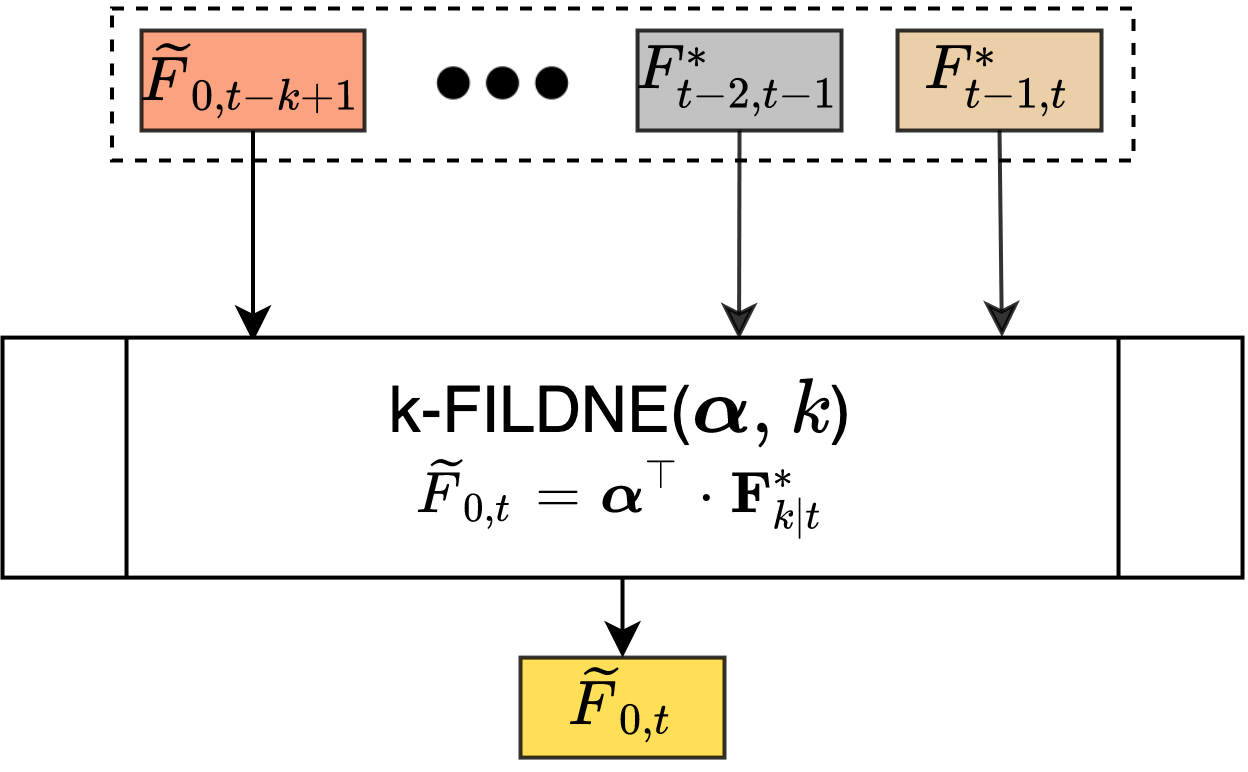}
        \caption{Single composition step of {\MethodNameGeneralized}. The algorithm obtains the embedding $\tilde{F}_{0,t}$ by computing the dot product between the parameter vector $\boldsymbol{\alpha}$ and the vector of $k$ aligned embeddings up to time $t$ $\boldsymbol{F}^*_{k|t}$.}
        \label{fig:icmen-composition}
    \end{subfigure}
    
    \caption{Overview of the proposed {\FrameworkName} algorithm in both variants.}
\end{figure*}

\paragraph{Step 1. Batch embedding}
For each new snapshot $G_{t-1, t} \equiv G_{t}$, we apply some network embedding technique -- called \emph{Base Method} -- and obtain $F_{t-1, t} \equiv F_t$ (see Figure \ref{fig:batch-embedding}). The choice of the Base Method is entirely up to the user. It can be an instance of Static Network Embedding (e.g. node2vec, LINE, DGI) and Temporal Network Embedding (e.g. CTDNE), both handling additional node/edge attributes if such are observed.

We focus on transductive Base methods, for which inferring representation for unseen examples is not possible in no other way than a full re-training of the model. One could also use inductive graph embedding methods (such as GraphSAGE or GCN-based approaches), but these do not suit well to our problem setting (our framework if fact provides a way to inductive learning itself).

\paragraph{Step 2. Alignment}
The random initialization and stochastic optimization adopted in embedding methods result in non-comparable realizations of representation vectors even for the same input data. The same applies when matching embeddings between two time steps, forcing us to run network embedding alignment before we go to the next step. The problem appears to be straightforward -- arranging embeddings in the form of matrices with matching rows lets us apply matrix alignment techniques. However, the nature of the problem to be solved should not be forgotten. Network evolution is not only about the appearance of new nodes but also about change in neighborhoods of existing ones. We are not interested in vertices, whose structural neighborhood topology was completely altered. Therefore, not all vectors should be considered when aligning embedding matrices. We introduce the concept of \emph{reference nodes} (later discussed in Section \ref{sec:reference_nodes}) or anchor nodes \cite{liu2016aligning} marked as $V_{\text{ref}}$. They are supposed to be (relatively) static over two neighboring batches. Having such a reference we can calculate the transformation matrix $Q$(as described in Definition \ref{def:matrix_alignment}) only on vectors representing those nodes and then apply in on the entire embedding matrix $F_{t}$ what results in $F^*_{t}$ (see Algorithm \ref{alg:embedding_aligment} and Figure \ref{fig:embedding-alignment}).

In case of the {\MethodNameGeneralized} method $F_{t}$ is aligned to the previous embedding $F^*_{t-1}$ which has been aligned in the former iteration. We store a vector of $k$ aligned embeddings $\boldsymbol{F}^*_{k|t}$.

\begin{algorithm}[ht]
\DontPrintSemicolon
\KwIn{$F_{t-1}, F_{t}$, $\{(a^{(v)}_{t-1}, a^{(v)}_{t}): v \in V_{t-1 \cap t}\}$, $\text{select}(\cdot, \cdot)$}

$S = \{s(a^{(v)}_{t-1}, a^{(v)}_{t}): v \in V_{t-1 \cap t} \}$ \Comment*[r]{Calculate node scores}
$V_{\textrm{ref}} = \text{select}(S, V_{t-1 \cap t})$ \Comment*[r]{Obtain reference nodes}
$U \Sigma W^\intercal = \text{SVD}(F_{t-1}(V_{\textrm{ref}})^\intercal F_{t}(V_{\textrm{ref}}))$\;
$Q = UW^\intercal$ \Comment*[r]{Compute transformation matrix}
\Return{$F_{t} Q$}
\caption{Embedding Alignment}
\label{alg:embedding_aligment}
\end{algorithm}

\paragraph{Step 3. Embedding composition}
In the {\MethodNameBasic} method (see Algorithm \ref{alg:basicicmen} and Figure \ref{fig:bi-composition}), at each iteration, we combine the previously composed embedding $\tilde{F}_{0, t-1}$ with its aligned embedding of the current snapshot $F^*_{t-1, t}$ using following \emph{convex combination}:

\begin{equation}
    \label{eq:basic-icmen-combine}
    \tilde{F}_{0, t} = \alpha \tilde{F}_{0, t-1} + (1 - \alpha) F^*_{t-1, t}
\end{equation}

At the beginning ($t = 2$) we use $\tilde{F}_{0, t-1} = F_{0, 1}$. The $\alpha$ parameter can be selected using search methods or experts' knowledge. Whenever $\alpha > 0.5$, it means that the past embedding is more important than the recent data increment in the Graph Stream.

\begin{algorithm}[ht]
\DontPrintSemicolon

\KwIn{$\tilde{F}_{0, t-1}$, $F_{t}$,
$\{(a^{(v)}_{t-1}, a^{(v)}_{t}): v \in V_{t-1 \cap t}\}$,
$\text{select}(\cdot, \cdot)$, $\alpha$}

${F^*_t = \text{Embedding Alignment}(\tilde{F}_{0, t-1}, F_{t}, a_{t-1}, a_t, \text{select})}$\; 
\vspace{-10pt}
$\tilde{F}_{0, t} = \alpha\tilde{F}_{0, t-1} + (1-\alpha)F^*_t$\;
\Return{$\tilde{F}_{0, t}$}

\caption{\MethodNameBasic}
\label{alg:basicicmen}
\end{algorithm}

The {\MethodNameGeneralized} version (see Algorithm \ref{alg:icmenmethod} and Figure \ref{fig:icmen-composition}) builds upon two parameters, that is $k \in \mathbb{N}$, the number of last embeddings combined by the algorithm, and $\boldsymbol{\alpha} \in \mathbb{R}^k$, a $k$-dimensional real-valued vector:
\begin{equation}
    \boldsymbol{\alpha} = [\alpha_1, ..., \alpha_k]^\intercal
\end{equation}
which is constrained to ${\{\boldsymbol{\alpha}: 0 \leq \alpha_i \leq 1, \sum_{i=1}^k \alpha_i = 1\}}$ namely the $k-1$-dimensional simplex. We propose a method to estimate $\boldsymbol{\alpha}$ what is described in Section \ref{sec:alpha_estimation}. Such construction enables us to combine more than two embeddings at once, in opposite to {\MethodNameBasic} scenario where we always combine 2 embeddings.

Let $\boldsymbol{F}^*_{k|t}$ denote the vector of $k$ embeddings up to time $t$:
\begin{equation}
\label{eq:calibrated_emebeddings_vector}
    \boldsymbol{F}^*_{k|t} =  [\tilde{F}_{0, t-k+1}, F^*_{t-k+2}, ..., F^*_{t}]^\intercal.
\end{equation}

The embedding rule is defined as follows:
\begin{equation}
    \label{eq:generalized-icmen-combination}
    \tilde{F}_{0, t} = \boldsymbol{\alpha}^\top \cdot \boldsymbol{F}^*_{k|t}
\end{equation}
which is the dot product of $\boldsymbol{\alpha}$ and the sequence of aligned embeddings $\boldsymbol{F}^*_{k|t}$.

\begin{algorithm}[ht]
\DontPrintSemicolon
\KwIn{$\boldsymbol{F}^*_{k-1 | t-1}$, $F_{t}$,
$\{(a^{(v)}_{t-1}, a^{(v)}_{t}): v \in V_{t-1 \cap t}\}$,
$\text{select}(\cdot, \cdot)$, $G_{t}$, prior}
${F^*_t = \text{Embedding Alignment}(F^*_{t-1}, F_{t}, a_{t-1}, a_t, \text{select})}$\; 
$\boldsymbol{F}^*_{k|t} = [\boldsymbol{F}^*_{k-1|t-1}; F^*_t]$ \;
$\boldsymbol{\alpha} = \text{Alpha Estimation}(\boldsymbol{F}^*_{k|t}, G_{t}, \text{prior}) $\;
 $\tilde{F}_{0,t}=\boldsymbol{\alpha}^\top \cdot \boldsymbol{F}^*_{k|t}$\;
\Return{$\tilde{F}_{0,t}$}
\caption{\MethodNameGeneralized}
\label{alg:icmenmethod}
\end{algorithm}

\subsection{Reference nodes selection}
\label{sec:reference_nodes}

To select appropriate reference nodes (see Figure \ref{fig:reference-nodes-selection}) we first introduce an \emph{activity function}
\begin{equation}
    a: V \rightarrow \mathbb{R}
\end{equation}
that for each node in the graph $G$ assigns a scalar describing its behaviour. In our experiments we use \emph{multi-degree} as the activity function. Activity is measured for common vertices $V_{t-1 \cap t}$ between two neighbouring snapshots.

The next step is to obtain a ranking of nodes best suited as a reference. To do so we apply a \emph{scoring function} -- in our case:
\begin{equation}
    s(a^{(v)}_{t-1}, a^{(v)}_t) = |a^{(v)}_{t-1} - a^{(v)}_t| \left( \frac{\pi}{2} - \arctan (\max \{a^{(v)}_{t-1}, a^{(v)}_t\}) \right),
\end{equation}
where $a^{(v)}_{t-1}$ and $a^{(v)}_t$ are $v$'s activities from neighboring snapshots. The resulting scores are \textbf{sorted in ascending order}. Finally, we are able to select a number of reference nodes based on the ranking. We propose the following schemes:
\begin{itemize}
    \item \textbf{Percent} -- based on the lowest score the top \textbf{p} percent of nodes is selected:
    $$select(S, V) = V_{ref} \subseteq \text{sort}_{S}(V), \ \text{s.t.} \ |V_{ref}| = \textbf{p}|V|$$
    
    \item \textbf{Multiplier} -- based on the lowest score the number of nodes is determined as \textbf{Md} (multiplier times the embedding size), but no more than $\textbf{p}_{\textbf{max}}$ (maximum percent) of common vertices:
    
    \begin{align*}
    \text{select}(S, V) = &V_{ref} \subseteq \text{sort}_{S}(V), \ \text{s.t.} \\
    &|V_{ref}| = \min(\textbf{Md}, \textbf{p}_{\textbf{max}}|V|)    
    \end{align*}
    
    \item \textbf{Threshold} -- all nodes with score lower than a given threshold \textbf{th} are selected as reference:
    $$\text{select}(S, V) = V_{ref} \subseteq V, \ \text{s.t.} \ \forall_{v \in V_{ref}} S^{(v)} \leq \textbf{th}$$
    
\end{itemize}

The methodology of determining reference nodes presented in this section is a generic solution. Other possible approaches include the usage of nodes' attributes or experts' knowledge.

\subsection{Alpha estimation}
\label{sec:alpha_estimation}

\begin{figure}[ht]
    \centering
    \includegraphics[width=\columnwidth]{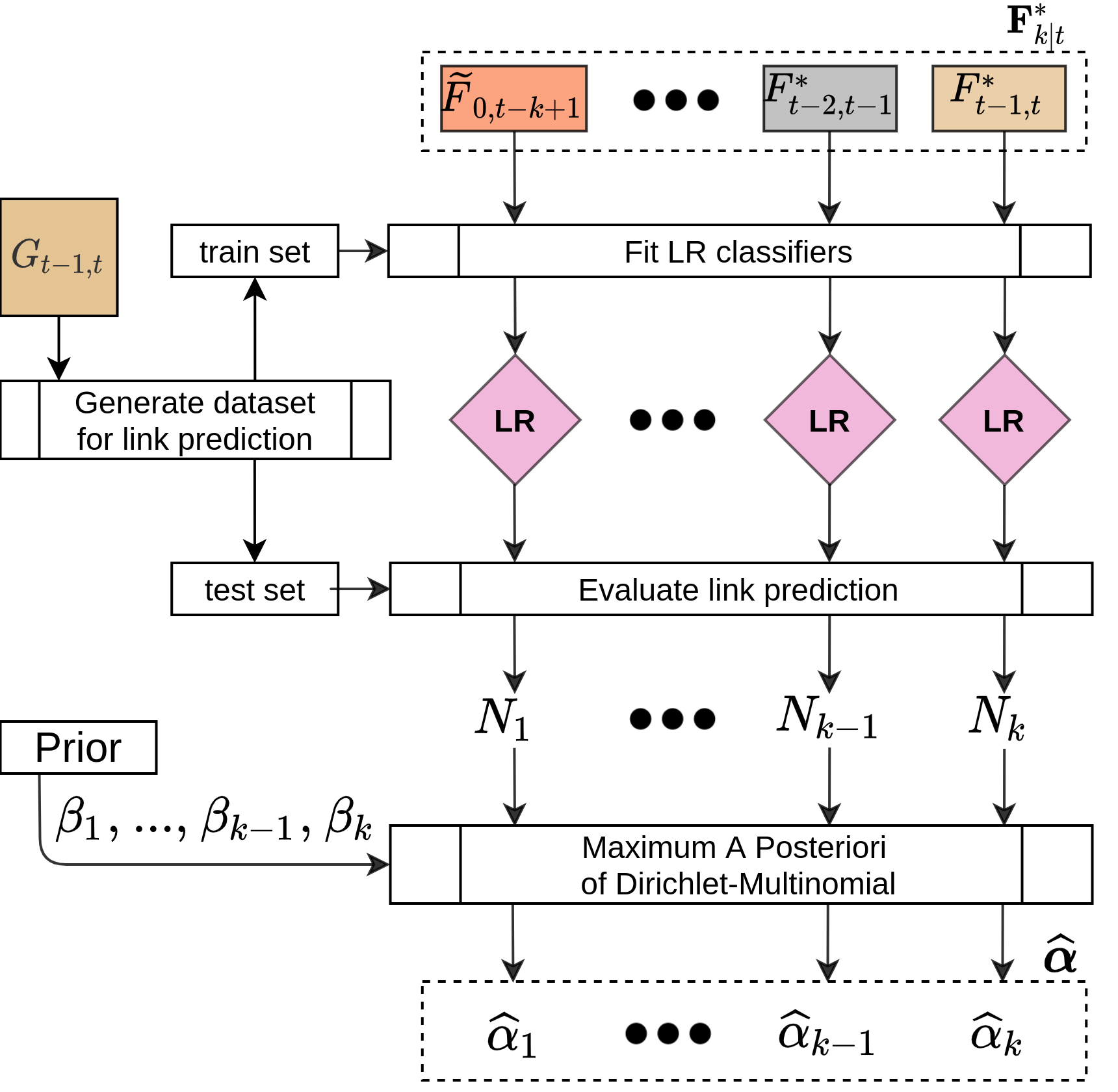}
    \caption{Alpha estimation. The newly arrived snapshot $G_{t-1,t}$ is prepared as train and test sets for link prediction. For each embedding in the vector $\boldsymbol{F}^*_{k|t}$ a logistic regression (LR) classifier is fitted on the train set and further evaluated on the test set. A poorly performing embedding should yield $\hat{\alpha}_i$ value that barely influence the final representation. This is achieved by Maximum A Posteriori estimation for the Dirichlet-Multinomial model fed with the classification results as described in Section \ref{sec:alpha_estimation}.}
    \label{fig:alpha-estimation}
\end{figure}

A significant problem arises while using the {\MethodNameGeneralized} model with $k$ parameters. Assuming that each parameter has the same number of considered values $|\Lambda|$, the model has a search space of size $O(|\Lambda|^k)$, i.e., it grows exponentially with each new dimension. Hence, we need to find a way to estimate the model parameters using an algorithm that is cheaper than a full search over the entire parameter space.
We propose an algorithm that uses Bayesian inference with assumption of Dirichlet-Multinomial distribution (see Algorithm \ref{alg:icmenalpha} and Figure \ref{fig:alpha-estimation}). We want to estimate the parameters $\hat{\boldsymbol{\alpha}} = \{\hat{\alpha}_1, \ldots, \hat{\alpha}_k\}$. For the prior we use the Dirichlet distribution:
\begin{equation}
    Dir(\boldsymbol{\alpha}|\boldsymbol{\beta}) = \frac{1}{B(\boldsymbol{\beta})} \prod_{i = 1}^{k} \alpha^{\beta_i - 1}_{i} \mathds{1}_{\{S_{k-1}\}}(\boldsymbol{\alpha}),
\end{equation}
where $B(\cdot)$ is the beta function used for normalization purposes. $\boldsymbol{\beta}$ reflects the prior knowledge about the distribution and $S_{k-1}$ denotes the $k-1$--dimensional simplex. We define two settings -- \emph{uniform}: ${\beta_1 = \beta_2 = \ldots = \beta_k = 1}$, where representations are equally important, and \emph{increasing}: ${\beta_1 < \beta_2 < \ldots < \beta_k}$, where more recent embeddings are assumed to be more significant.

The likelihood function has the form:
\begin{equation}
    p(\boldsymbol{D}|\boldsymbol{\alpha}) = \prod_{i = 1}^{k} \alpha^{N_i}_{i}
\end{equation}
where $\boldsymbol{D} = [N_1, N_2, ..., N_k]$ is the vector of class occurrences from the link prediction experiment described below. Note that $N = \sum_{i=1}^k N_i$ is the total size of the sample.

Each of the embeddings $\tilde{F}_{0, t-k+1}$, $F^*_{t-k+2}$, $\ldots$, $F^*_{t}$ (see Equation \ref{eq:calibrated_emebeddings_vector}) is a separate class in the Multinomial distribution. We take edges from the most recent snapshot $G_{t}$ and split them into the train and test sets. Negative edges are sampled in both groups in numbers enabling class balance. We fit a set of Logistic Regression classifiers with input vectors for each edge, built as Hadamard product of node embeddings, and the outputs denoting edge existence. If a link was correctly predicted with several embeddings, we randomly choose only a single representation. If none of them can provide the correct classification of the link, such an edge is removed from the sample. The class counts (correct predictions) $[N_1, N_2, ..., N_k]$ are measured on the test set.

Using the above assumptions we estimate the parameters $\boldsymbol{\hat{\alpha}}$ as the maximum a posteriori probability (MAP) of the Dirichlet-Multinomial model \cite{Murphy2012}:
\begin{equation}
    \hat{\alpha}_j = \frac{N_j + \beta_j - 1}{N + \sum_{i=1}^k \beta_i - k}
\end{equation}

\begin{algorithm}[ht]
\DontPrintSemicolon
\KwIn{$\boldsymbol{F}^*_{k|t}, G_{t}, \text{prior}$}
Generate link prediction dataset over graph $G_{t}$\;
Fit Logistic Regression classifiers on train set\;
Evaluate link prediction on test set and report correct predictions $[N_1, N_2, \ldots, N_k]$\;
Set $\boldsymbol{\beta}$ according to $prior$ distribution\;
$\displaystyle \hat{\boldsymbol{\alpha}} = \left[\frac{N_1 + \beta_1 - 1}{N + \sum_{i=1}^k \beta_i - k}, \ldots, \frac{N_k + \beta_k - 1}{N + \sum_{i=1}^k \beta_i - k}\right]$\;
\Return{$\hat{\boldsymbol{\alpha}}$}
\caption{Alpha Estimation}
\label{alg:icmenalpha}
\end{algorithm}

\subsection{Missing embeddings: new and inactive nodes}

In the {\MethodNameBasic} algorithm, the way we establish the embedding for new or disappearing nodes is trivial. In such a case at time $t$, we only have one out of two embeddings, and Eq. \ref{eq:basic-icmen-combine} cannot be applied. For new (previously unseen) nodes $v$, we use the its embedding from the most recent snapshot as the estimated embedding, i.e. $\tilde{F}_{0, t}(v) = F^*_{t-1, t}(v)$. Contrary, if a node does not appear in most recent graph snapshot, we use its previously estimated embedding, i.e. $\tilde{F}_{0, t}(v) = F^*_{0, t-1}(v)$.

For the {\MethodNameGeneralized} method, the problem is more complex. The number of input embeddings $k'$ varies between $1$ and $k$. If there are all embeddings available, we simply apply Eq. \ref{eq:generalized-icmen-combination}, and if $k'=1$, then we proceed with the same idea as for the {\MethodNameBasic} approach. 
Otherwise, we take the estimated $\hat{\boldsymbol{\alpha}}$ coefficients (see Section \ref{sec:alpha_estimation}) corresponding to all $k'$ available embeddings for a given node. Next, we normalize those values, so that they sum up to $1$. Then, we apply Eq. \ref{eq:generalized-icmen-combination} assuming that we only have $k'$ components.

\subsection{Complexity analysis}

To estimate the time and space complexity of our algorithm, we consider the scenario of a single incremental step, i.e. the moment when we have $k - 1$ past embeddings $[\tilde{F}_{0, t-k+1}, F^*_{t-k+2}, ..., F^*_{t-1}]^\intercal$ and a new graph snapshot $G_{t}$ appears. We use a random walk based embedding algorithm (Node2vec) and the {\MethodNameGeneralized} version. We denote $O_T(\cdot)$ and $O_S(\cdot)$ as the time and space complexities, respectively. Considering each algorithm step separately, we get:

\begin{enumerate}
    \item[(1)] First, we run Node2vec: 
    $$O_T(\text{Node2vec}) = O(d|V_{t-1, t}|)$$
    $$O_S(\text{Node2vec}) = O(\gamma \omega |E_{t-1, t}|),$$
    where $\gamma$, $\omega$ are the number and length of random walks respectively.
    
    \item[(2)] We obtain reference nodes by sorting all vertices $V_{t-1 \cap t}$ by their activity score; it takes: 
    $$O_T(\text{sort}) = O(|V_{t-1 \cap t}| log|V_{t-1 \cap t}|)$$
    $$O_S(\text{sort}) = O(|V_{t-1 \cap t}|)$$
    Next, we solve the Orthogonal Procrustes (OP) problem during embedding alignment: 
    $$O_T(\text{OP}) = O(d^3)$$,
    $$O_S(\text{OP}) = O(d^2)$$.
    
    \item[(3a)] Link prediction requires sampling of non-existing edges (in the same number as existing edges; Negative Sampling) takes: 
    $$O_T(\text{NS}) = O_S(\text{NS}) = O(|E_{t-1, t}|)$$
    We train Logistic Regression using all $k$ embeddings: 
    $$O_T(\text{LR}) =  O(kd^2|E_{t-1, t}|)$$
    $$O_S(\text{LR}) = O(kd|E_{t-1, t}|)$$
    Finally we apply LR and aggregate the Class Counts, which takes:
    $$O_T(\text{CC}) = O(k * |E_{t-1, t}|)$$
    $$O_S(\text{CC}) = O(k)$$
    
    \item[(3b)] The estimation of $\boldsymbol{\alpha}$ using Dirichlet-Multinomial model is done by simple division of $k$ numbers, so it takes: 
    $$O_T(\text{dirichlet}) = O_S(\text{dirichlet}) = O(k)$$
    
    \item[(4)] for the composition of all embeddings, we perform a Weighted Matrix Addition, which takes: 
    $$O_T(\text{WMA}) = O_S(\text{WMA}) = $$
    $$ = O(d|V_{0,1}| + \ldots + d|V_{t-1, t}|) = O(d|V|)$$
    where $|V|$ is the number of vertices in the whole graph.
\end{enumerate}

Therefore, the time and memory complexities of {\FrameworkName} equals the sum of all above steps' complexities. The size of the last snapshot $V_{t-1, t}, E_{t-1, t}$ is smaller than of the whole graph $V_{0, t}, E_{0, t}$: $|V_{t-1, t}| < |V_{0, t}| = |V|$, $|E_{t-1, t}| < |E_{0, t}| = |E|$. Let us note that the $d$ and $k$ hyper-parameters do not scale with the size of the network. Hence, we can approximate the total {\FrameworkName} complexity to: $O_T({\FrameworkName}) \approx O(|V| \log |V| + |E|)$ and $O_S({\FrameworkName}) \approx O(|V| + |E|)$. 

\subsection{{\FrameworkName} versus other Dynamic Graph Embedding methods}
\label{sec:methods_comparison}
Let us now consider the differences between different approaches of dynamic embedding presented in the literature. First off, we aim at highlighting the differences in data requirements of each approach while computing embedding for new snapshot. The most unfavorable group of methods requires to store whole graph information from the very beginning to time $t$, i.e. $G_{0, t}$. These are \textbf{CTDNE}, \textbf{Online-Node2Vec}, \textbf{tNodeEmbed}, \textbf{dyngraph2vec}, \textbf{Dynnode2vec}. Additionally, tNodeEmbed method requires all intermediate embeddings  $F_{0, 1}, \ldots, F_{0, t-1}$. Another group of methods necessitate only the most recent graph snapshots, i.e. $G_{t-1}, G_{t}$ (\textbf{GloDyne}). 
Our approach, in opposite to all previous, requires only one embedding $\tilde{F}_{0, t-1}$ (\textbf{\MethodNameBasic}) or $k$ embeddings $\tilde{F}_{0, t-k+1}, \ldots, F^*_{t-2}, F^*_{t-1, t}$ (\textbf{\MethodNameGeneralized}) and the activity of nodes from $G_{t-1}$, i.e. $a^{(v)}_{t-1} \forall v \in G_{t-1}$. 

Second off, we want to emphasize the difference in the easiness of methods' hyperparameters tuning. Such dynamic network embedding methods as \textbf{dyngraph2vec}, \textbf{tNodeEmbed} 
strongly rely on deep network architecture and all related optimization problems. \textbf{Online-Node2Vec} requires to specify time-dependent hyperparameters that are not very intuitive: time-decay and half-life, which must be set by an expert or through extensive searching. Our approach limits the number of hyperparameters to only two simple and intuitive: the number of reference nodes, combination weight $\alpha$ ({\MethodNameBasic}), or \textit{prior} ({\MethodNameGeneralized}). 

Most importantly, third off, our approach allows utilizing embeddings of choice (methods or already calculated vectors) in contrast to all other methods, which rely on specific embedding objectives. It enables our method to use additional network-related data, e.g., nodes' attributes (using, e.g., \textbf{DGI}).

The \textbf{LCF} method proposed in \cite{trivedi2018structural} does not fit to any of the above-mentioned criteria. The authors propose a similar framework to our {\FrameworkName} method. However, there are some non-negligible differences. In contrast to them, we employ a convex combination, a special case of the linear combination. Such an approach allows us to keep a nearly constant order of magnitude of embeddings vectors, whereas for linear combination with all coefficients equal to one (as proposed in \cite{trivedi2018structural}), the magnitude depends on the number of combined embeddings. Further, they introduce an exponential decay based approach where the most recent representation prevails the final embedding. Such an assumption is not flexible, whereas, in our proposed {\FrameworkName} methods, the combination parameters $\alpha$ are dynamically estimated from the data. Moreover, our experiments show that the assumption of exponential decay is not satisfied in the real world temporal networks (See Fig. \ref{fig:basic-icmen-sensitivity}). The authors do not neglect the fact that proximity-based embeddings are not comparable across timestamps and therefore apply embedding alignment. However, they do it naively by taking all the available reference nodes between snapshots. Our approach measures the stability of nodes' activities to select the most appropriate ones. Finally, the reported results are not compared to the other state-of-the-art Dynamic Graph Embedding methods. Furthermore, not commonly used accuracy metrics makes it impossible to compare with them directly.

\section{Experiments}
\label{sec:experiments}
We evaluate the performance of our proposed algorithm in several experiments and compare the results against commonly used baseline methods as well as the Base methods in naive Dynamic Graph Embedding setting. Firstly, we train all methods and use the computed embeddings in a \emph{link prediction} experiment, which should check whether these vector representations encode structural graph information, which can be used to distinguish connected nodes from non-connected ones. Next, for two datasets with edge labels, we use the same embeddings to perform \emph{edge classification} -- we measure the classification quality. We would like to preserve the distances between nodes in the resulting embedding space when learning representations on graphs. Hence, we compute metrics used in \emph{graph reconstruction} (distortion, mAP). Finally, we perform a hyperparameter sensitivity study to show how they influence the model's performance.

\subsection{Datasets}
\label{sec:datasets}
We conduct experiments on seven popular dynamic graph datasets downloaded from the Network Repository \cite{network_repository} -- we summarize their statistics in Table \ref{tab:datasets}. Selected datasets vary in the total time span between the first and last event (edges): from 2 days (\textit{hypertext09}) up to about 5 years (\textit{bitcoin-otc}). Three of them contain directed edges. Each dataset is used in link prediction and graph reconstruction tasks. Availability of edge labels in \textit{bitcoin-alpha} and \textit{bitcoin-otc} datasets allows us to perform edge classification. In preliminary experiments, we have checked several sizes of node embeddings for all networks and selected the best ones for each one (see Table \ref{tab:datasets}). We note the differences in the characteristics of selected datasets. The two bitcoin networks exhibit a meager amount of intersecting nodes and edges across pairs of consecutive snapshots when compared to the others. The fb-forum, fb-messages, and hypertext09 graphs stay at a moderate level of variability, while enron-employees and radoslaw-email seem stable.
\begin{table*}[ht!]
    \caption{Statistics of graph datasets. $\mathbf{|V|}$ - no. of nodes, $\mathbf{|E|}$ - no. of temporal edges (events), $\mathbf{D}$ - density of the graph, \textbf{Directed} - is the graph  directed or not. \textbf{LP, EC, GR} - dataset was used in \textbf{L}ink \textbf{P}rediction, \textbf{E}dge \textbf{C}lassification, \textbf{G}raph \textbf{R}econstruction tasks, respectively. \textbf{Avg. Jaccard Index} of nodes and edges computed as the mean of respective Jaccard Indexes across all consecutive snapshot pairs.}
    \label{tab:datasets}
    \begin{center}
        \begin{small}
            \begin{sc}
                \scalebox{0.9}{
                    \begin{tabular}{l|rrrrccccc}
                        \toprule
                        Dataset & $|V|$ & $|E|$ & $D$ & Timespan & Directed & Tasks & Embedding & \multicolumn{2}{c}{Avg. Jaccard Index}\\
                        &&&& (days)&&& size & nodes & edges\\
                        \midrule
                        enron-employees & 151 & 50 572 & 4.466 & 1 138 & $\times$ & LP, GR & 32 & 0.864 & 0.361\\
                        hypertext09 & 113 & 20 818 & 3.290 & 2 & $\times$ & LP, GR & 32 & 0.816 & 0.142\\
                        radoslaw-email & 167 & 82 927 & 2.991 & 271 & $\surd$ & LP, GR & 32 & 0.903 & 0.430\\
                        fb-forum & 899 & 33 720 & 0.084 & 164 & $\times$ & LP, GR & 128 & 0.692 & 0.253\\
                        fb-messages & 1 899 & 61 734 & 0.034 & 216 & $\times$ & LP, GR & 128 & 0.516 & 0.093\\
                        bitcoin-alpha & 3 783 & 24 186 & 0.002 & 1 901 & $\surd$ & EC, LP, GR & 32 & 0.217 & 0\\
                        bitcoin-otc & 5 881 & 35 592 & 0.001& 1 903 & $\surd$ & EC, LP, GR & 128 & 0.213 & 0\\
                        \bottomrule
                    \end{tabular}
                }
            \end{sc}
        \end{small}
    \end{center}
\end{table*}

\subsection{Experimental setup}
    In the following paragraphs, we explain how we have configured the experimental environment.
    
    \paragraph{Base methods}
    We evaluate several state-of-the-art and representative node embedding algorithms from different method families, i.e. based on: random walks (\textbf{Node2vec} \cite{grover2016node2vec}, \textbf{CTDNE} \cite{Nguyen2018}), matrix factorization (\textbf{HOPE} \cite{Ou2016}, \textbf{LLE} \cite{roweis2000nonlinear}, \textbf{LE} \cite{belkin2003laplacian}), graph neural networks (\textbf{DGI} \cite{velickovic2019deep}) and general function optimization (\textbf{LINE} \cite{tang2015line}). 
    
    These methods do not provide the ability to perform training incrementally. We use these methods to compute two kinds of embeddings: 
    \begin{itemize}
        \item baseline embeddings to compare our proposed {\FrameworkName} method and other streaming/incremental ones against, i.e., the node representations are computed on the cumulative snapshots $G_{0, t}$ at a given time $t$;
        \item embeddings for non-overlapping graphs $G_{t, t+1}$, which are further consumed by our proposed {\FrameworkName} method.
    \end{itemize}
    
    \paragraph{Streaming methods}
    In terms of other incremental methods designed for graph streams embedding, we compare our method to: \textbf{tNodeEmbed} \cite{singer2019node}, two variants of \textbf{Online-Node2vec} \cite{lee2019dynamic} (\textbf{StreamWalk, SecondOrder}) and \textbf{dyngraph2vec} \cite{goyal2020dyngraph2vec} (in the \textbf{AE-RNN} version).
    
    \paragraph{Convergence issues}
    During the experiments, we found out that the LLE \cite{roweis2000nonlinear} method did not converge on three datasets (fb-messages, bitcoin-alpha, bitcoin-otc). We checked several hyperparameter settings of the underlying optimizer, but none of them fixed this issue. We decided not to report the results of this method for these datasets.
    
    \paragraph{Data preprocessing}
    For each dataset, we apply the following preprocessing steps: (1) we take all edges and sort them by time in an ascending manner, (2) we split these edges into ten equally sized parts according to time, (3) we convert each edges chunk into a graph snapshot and call it $G_{t, t+1}$, where $t \in \{0, \ldots, 9\}$. In total, we obtain \emph{10 non-overlapping graph snapshots}. We also save cumulative graphs which accumulate all edges from the beginning, i.e. $G_{0, t}$, which contain all edges from snapshots $G_{0,1}, G_{1,2}, \ldots, G_{t-1, t}$.
    
    Note that tNodeEmbed updates its internal model for every single timestamp. For the method to be comparable in our experimental scenario, we assume that each of the ten snapshots, as mentioned earlier, is equivalent to a single timestamp that is processed by tNodeEmbed. For this method, we also reuse already calculated Node2vec embeddings. 
    
    \paragraph{Embedding calculation}
    We train all of the above-mentioned methods and obtain two groups of embeddings: (1) \textit{cumulative} $F_{0, t}$ for $t \in \{0, \ldots, 9\}$, which are computed in all $G_{0, t}$ using the \textit{Base} and \textit{Streaming methods}; (2) \textit{non-overlapping} $F_{t, t+1}$ using the Base methods and further combined by our proposed {\FrameworkName} methods to obtain \textit{cumulative} embeddings $\tilde{F}_{0, t}$. Unsupervised embeddings (all but tNodeEmbed) are trained once and used in all downstream tasks. tNodeEmbed is a supervised method, and the embeddings are trained for link prediction and edge classification individually. We repeat the training and evaluation procedure 30 times, reporting averaged statistics, to address the random initialization and stochastic optimization used in the methods (e.g., random walk generation in Node2vec and CTDNE).
    
    \paragraph{Mean ranks}
    For each snapshot and dataset, we establish a ranking of methods based on the average of 30 runs. We summarise it as \emph{mean rank}, which is the average ranking of methods. We report this score in each results table in the form of a separate column. Based on this score, we mark the three best methods in bold.
    
    \paragraph{Fine-tuning}
    To provide a fair comparison of all methods, we decided to perform a hyperparameter search, with an equal number of 100 iterations, overall methods. We use Tree of Parzen Estimators (TPE) \cite{bergstra2011algorithms} (implemented in the HyperOPT \cite{bergstra2013making} package) choosing most appropriate hyperparameters.
    
    \paragraph{Reproducibility}
    To allow other researchers and developers to try out our proposed {\FrameworkName} model, we make our code available at \url{https://gitlab.com/fildne/fildne}. We also publish all experiments in the form of a DVC pipeline \cite{dvc}, so they can be easily reproduced.
    
\begin{figure}[H]
    \centering
    \includegraphics[width=\columnwidth]{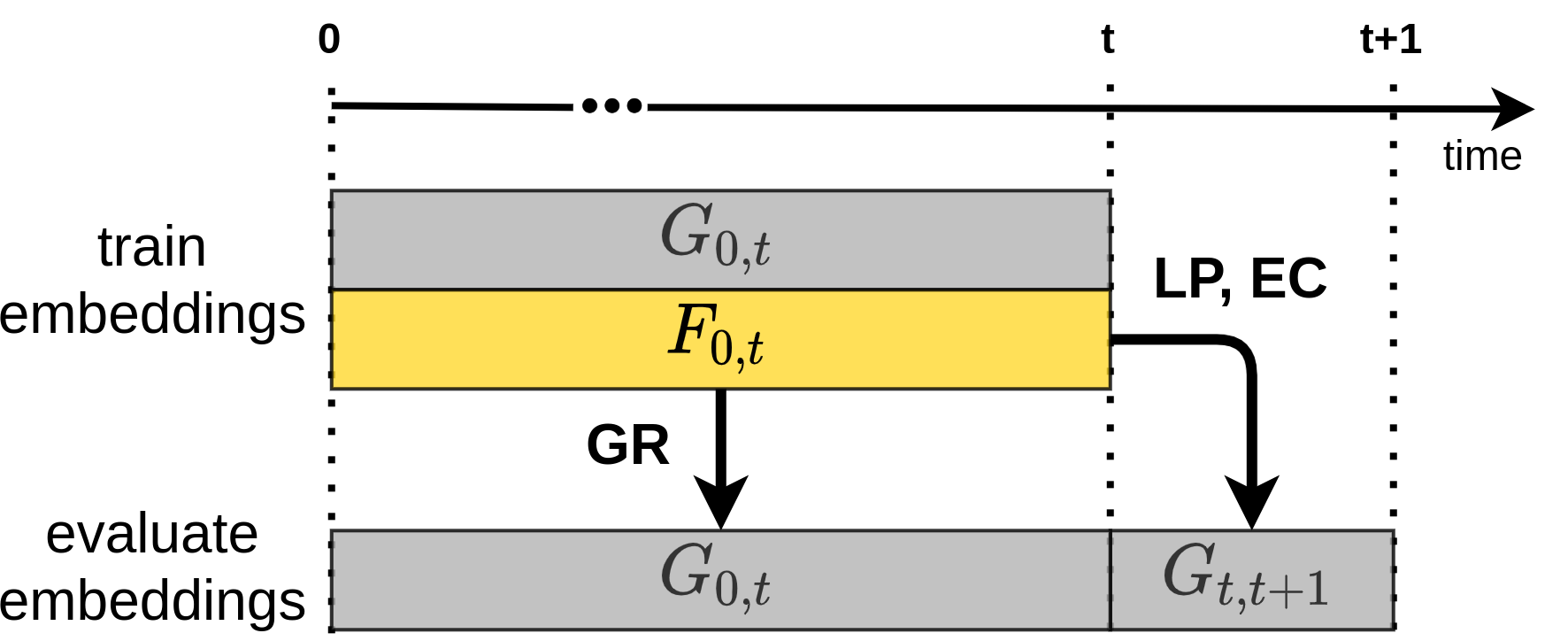}
    \caption{Evaluation protocol. Link prediction (LP) and edge classification (EC) tasks are evaluated on the \textbf{next} snapshot $G_{t, t+1}$, whereas graph reconstruction (GR) task is evaluated on the \textbf{corresponding} graph snapshot $G_{0,t}$.}
    \label{fig:task-setup}
\end{figure}    
    
\subsection{Link prediction\label{sec:link-prediction}}
    \paragraph{Setup}
    For each graph snapshot $G_{t, t+1}$, we generate a link prediction dataset. We split existing edges into a train (75\%) and test (25\%) dataset and mark as class 1. Then we sample the same number of non-existing edges from the graph (negative samples -- class 0).
    
    We implement \textbf{next snapshot prediction} evaluation scheme. The embedding methods are trained on cumulative snapshots, i.e. $G_{0,1}, G_{0,2}, \ldots, G_{0, t}$ and on non-overlapping ones, i.e. $G_{0,1}, G_{1,2}, \ldots, G_{t-1, t}$. The latter embeddings are further combined by our proposed {\FrameworkName} method to obtain $\tilde{F}_{0, t}$). For each embedding $F_{0, t}$, we evaluate it on the \textbf{next} snapshot $G_{t, t+1}$ (see Figure \ref{fig:task-setup}). We follow well established protocol to provide link-prediction by means of classification if an edge exists \cite{grover2016node2vec}. Thus, we combine pair of node embeddings into edge features using \textbf{Hadamard} operator and feed \textbf{logistic regression} classifier. 
    
    Note that the Base methods do not provide any mechanism for obtaining embeddings for previously unseen nodes. We modify the sampling procedures mentioned above to consider edges (and negative edges) with nodes present in the appropriate embedding matrix.
    
    \paragraph{Results}
    Experiments on the cumulative snapshots with Base methods are presented in Table \ref{tab:base-method-selection}. We see that LINE and Node2vec achieve the best results on all datasets. Based on the mean rank, we observe that the \textbf{3 best methods (LINE, Node2vec, CTDNE)} outperform other methods. Hence, in the other part of this article, we will focus only on these -- using them as Base methods for our proposed {\FrameworkName} algorithm. Although, if the reader is interested in full results, we provide those in the Appendix \ref{app:full-tables-mean}.
    
    \input{tables_new/lp/base-methods-selection-lp}

    Based on the mean rank, we see in Table \ref{tab:mean-auc-lp} that among the three best methods, there are two embeddings composed with {\FrameworkName}. The results clearly show that our method beats other streaming approaches in all cases besides the bitcoin-alpha dataset. Considering the three best methods, we observe that for fb-messages, enron-employees, and hypertext09, there are two {\FrameworkName}-based representations, whereas for bitcoin-otc and radoslaw-email our proposed method is placed in all three of them.
    
    We notice that in the LINE Base method on bitcoin-otc and radoslaw-email, both our {\FrameworkName} methods result in high-performance gain compared to the vanilla Base algorithm. For other methods and datasets, we see comparable results. In general, we see that the gap between {\MethodNameGeneralized} and {\MethodNameBasic} is small.

    \input{tables_new/lp/mean-auc-lp}

\subsection{Edge classification}
    \paragraph{Setup}
    This downstream task is defined similarly to the link prediction setup (see Figure \ref{fig:task-setup}); however, the dataset is built differently. We do not need to sample negative instances. We use the bitcoin-alpha and bitcoin-otc graphs, where each edge has a label assigned (besides the timestamp) -- it represents the trust value of a transaction (values range from -10 to 10). We choose a threshold of 0 to define two classes (values below 0 -- class 0, negative trust, untrusted; values equal or higher than 0 -- class 1, positive trust). We use the same edge embeddings as in the link prediction task (all methods but tNodeEmbed, which is retrained in a supervised way for this problem explicitly) to train a logistic regression classifier. As the resulting datasets are mostly imbalanced, we set the "class\_weight" argument to "balanced" to let the algorithm automatically determine appropriate class weights (we use the Scikit-learn implementation).
    
    \paragraph{Results}
    We report results of edge classification in Table \ref{tab:mean-auc-ec}. The tNodeEmbed method outperforms others, but it is re-fitted in the same task, while the others reuse previously trained embeddings. Nevertheless, among the three best results based on the mean rank, two are obtained by {\FrameworkName}. For the bitcoin-otc dataset, we see that {\FrameworkName} improves AUC of edge classification for all Base methods. Analogously to link prediction (see Section \ref{sec:link-prediction}), we see that {\MethodNameBasic} and {\MethodNameGeneralized} perform similarly as well as our method beats other streaming approaches (except tNodeEmbed).
    
    \input{tables_new/ec/mean-auc-ec}

\subsection{Graph reconstruction}
    \paragraph{Setup}
    Contrary to link prediction and edge classification tasks, we \emph{do not use} here the next snapshot prediction (see Figure \ref{fig:task-setup}). The main goal in graph reconstruction problems is to tell how well the embedding represents the graph it was trained on, i.e. a given embedding $F_{0, t}$ is evaluated on its corresponding graph snapshot $G_{0,t}$ (in case of {\FrameworkName} we check how well the composed Base method embeddings reflect the original cumulative graph). Graphs are transformed to static ones in order to fulfill the graph reconstruction framework. We use the two metrics \cite{de2018representation}:
    \begin{itemize}
        \item a local one -- \textbf{mAP (mean Average Precision)}, which captures local graph properties; for any node and its embedding vector it checks how many of the nearest vectors (in the sense of euclidean norm) in the embedding space are actually first-order neighbours of this node:
        \begin{equation}
            mAP = \frac{1}{|V|}\sum_{v \in V}{\frac{1}{\text{deg}(v)}}\sum_{i=1}^{|N_v|}{\frac{|N_v \cap R_{v, w_i}|}{|R_{v, w_i}|}},
        \end{equation}
        where $\text{deg}(v)$ denotes the degree of $v$, $N_v = \{w_1$, $w_2$, $\ldots$, $w_\text{deg}(v)\}$ -- neighborhood of $v$, $R_{v, w_i}$ -- is the smallest set of such points that contains $w_i$ (that is, $R_{v, w_i}$ is the smallest set of nearest points in the embedding space required to retrieve the $i$th neighbor of $v$).
        
        \item a global one -- \textbf{distortion}, which compares the distances in the embedding space (euclidean norm) with the distances in the graph (shortest path lengths), the embedding distances are normalized to be within the range $[1; n]$, where $n$ is the longest shortest path length (also called \emph{diameter} of the graph).
        \begin{equation}
            D = \frac{1}{\binom{n}{2}}\left(\sum_{u, v \in U: u \neq v}{\frac{|d_E(u, v) - d_G(u, v)|}{d_G(u, v)}}\right),
        \end{equation}
        where $n$ is the number of nodes, $d_G$ are the graph distances and $d_E$ are the embedding distances.  
    \end{itemize}
    
    Note that we are interested in higher mAP values and lower distortion values (with 100\% and 0 as the ideal values, respectively). We also do not need an auxiliary classifier, like logistic regression for link prediction and edge classification. Using a given graph and its respective embedding, we compute the metrics.
    
    \paragraph{Results}
    Table \ref{tab:mean-gr-map} summarizes the graph reconstruction task results as the mean Average Precision. Considering the mean ranks, we see both {\MethodNameBasic} and {\MethodNameGeneralized} models in the top three methods. One of them ({\MethodNameGeneralized} with Node2vec embeddings) achieves the best results for fb-forum, improving the pure Node2vec by approximately $10\%$ percentage points. Overall, Node2vec significantly outperforms other methods on all the remaining datasets. We observe that the other streaming approaches perform poorly -- they allocate themselves at the three last positions in the mean ranks.
    
    \input{tables_new/gr-mAP/mean-gr-mAP}

    We also examine distortion as a graph reconstruction measure (see Table \ref{tab:mean-gr-distortion}) for which our proposed method improves the results of Base methods in most cases, or it stays at a comparable level. Contrary to mAP, the competitive streaming methods occupy two of the top three positions alongside {\FrameworkName}. Moreover, our method achieves the best performance on fb-forum and hypertext09 datasets.

    \input{tables_new/gr-distortion/mean-gr-distortion}

\subsection{Time and memory costs}
    \paragraph{Setup}
    In this experiment, we measured the time of computing embeddings using all of the considered methods, i.e.:
    \begin{itemize}
        \item for Base and streaming methods it is only the time needed to compute the embedding of the graph $G_{0,t}$ (either in batched or streaming manner);
        \item for both {\MethodNameBasic} and {\MethodNameGeneralized} we sum up the time of computing: node activities in graph $G_{t-1, t}$, embedding $F_{t-1,t}$ using a given Base method, the calibration procedure of this new embedding to previous ones, alpha estimation (for the {\MethodNameGeneralized}) and the time of applying the {\FrameworkName} composition equation.
    \end{itemize}
    We also measure the peak (highest) memory consumption of the procedures mentioned above. Note that this also includes the reading of the graphs into memory, as well as previously saved models (for streaming methods) or calibrated embeddings and node activities (for both {\FrameworkName} methods).
    
    The time measurements are performed while computing embedding in the main pipeline, so we have 30 measurements for each scenario. Meanwhile, the memory measurements are done in a separate branch of the pipeline, as it requires to probe the current memory usage with a relatively high frequency to obtain accurate results. It leads to a massive slowdown of the embedding algorithms, so eventually, we decided to perform five repetitions of these measurements.
    
    For the tNodeEmbed method, we report calculation time and memory utilization measured on the link prediction task.
    
    \paragraph{Results}
    From the time measurements reported in Table \ref{tab:mean-calculation-time}, it is visible that our proposed {\FrameworkName} method is the fastest for all datasets. The speedup ratio between the fastest {\FrameworkName} configuration and the next fastest competitive method ranges from approximately two to three times. Moreover, {\MethodNameGeneralized} is slower than the Basic version, due to the alpha estimation procedure, described in Section \ref{sec:alpha_estimation}. However, the actual time difference is not significant -- in most cases, {\MethodNameGeneralized} is about $1$ second slower, what corresponds to a slowdown of $1\% - 10\%$.
    In the case of memory utilization (see Table \ref{tab:mean-mm}), two {\FrameworkName} configurations are on average in the top three best-performing methods. It does not mean that our method always improves the Base method, especially on LINE, we can see no progress nor degradation. The rise of memory consumption is present for the hypertext09 dataset due to the small size of the network. We see that the Online-Node2vec methods are well optimized for memory, contrary to other streaming methods.
    
    \input{tables_new/calculation-time/mean-calculation-time}

    \input{tables_new/mm/mean-mm}

\subsection{Gain results}

\subsubsection{Comparison to Base methods\label{sec:gain-loss-base-methods}}

\begin{figure*}[!ht]
    \centering
    \includegraphics[width=\textwidth]{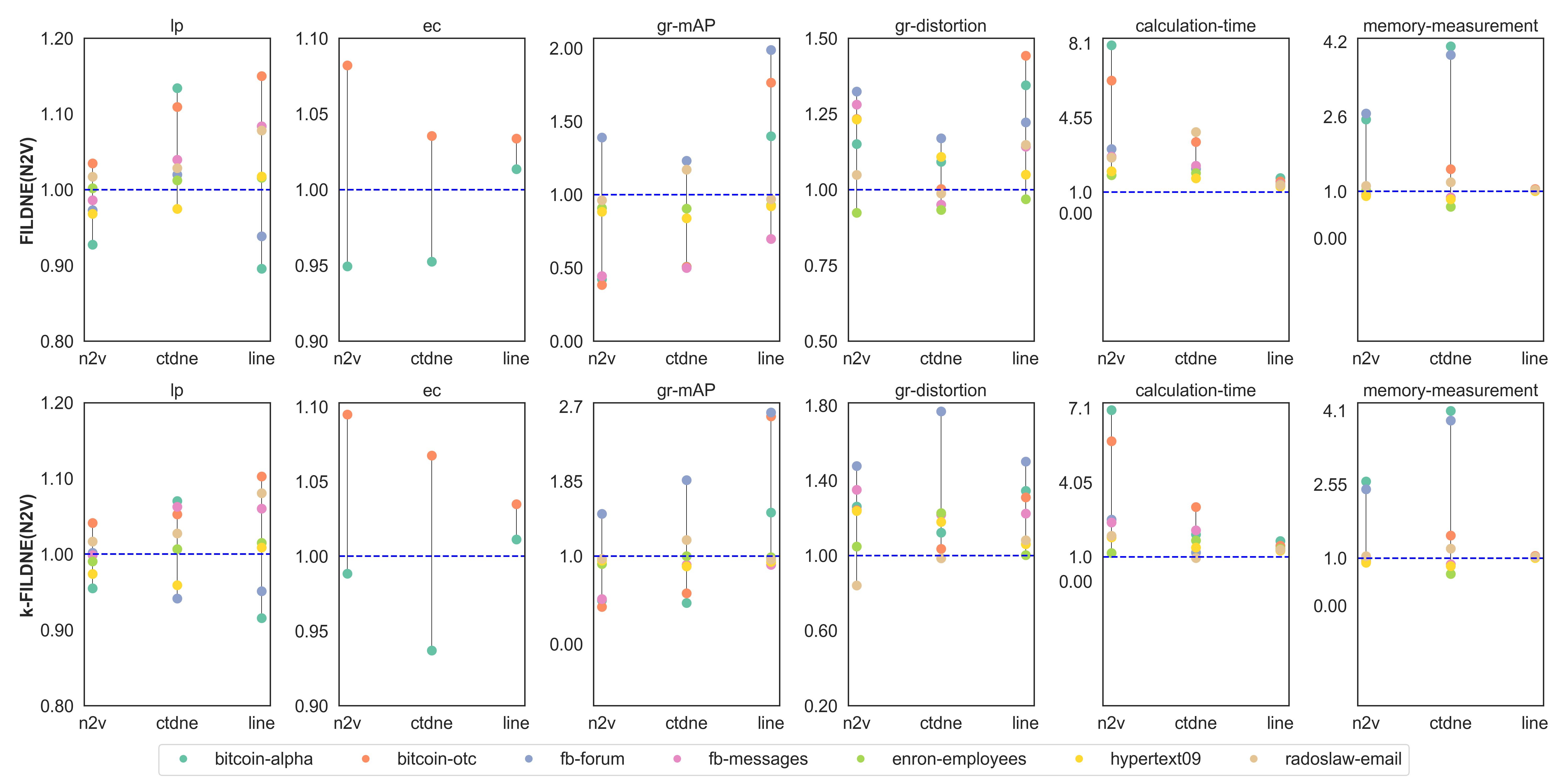}
    \caption{Mean gain scores for {\MethodNameBasic} and {\MethodNameGeneralized} compared to corresponding Base methods. Points above the dashed line (score equal to $1$) indicate a performance gain, whereas points below indicate loss.}
    \label{fig:gain-loss-base}
\end{figure*}

For all of the experiments, we compare {\FrameworkName} results to the corresponding Base method, computing \emph{gain scores} interpreted as a gain whenever above $1$, and as a loss whenever below. To calculate these scores first, we average all 30 retrains of each method for each snapshot. Next, we divide the results of {\FrameworkName} by the ones obtained with the corresponding Base method. We compute the mean of those ratios and report it as the gain score (see Figure \ref{fig:gain-loss-base}). Note that we are interested in lower values for distortion, time, and memory consumption, so we take the fraction's reciprocal. Each pair (dataset, Base method) can be viewed from the perspective of link prediction performance, edge classification performance, graph reconstruction quality, memory consumption, and time required for computation. Such a perspective allows for a thorough inspection of our method's properties evaluated on real-world datasets.

In the case of link prediction, edge classification, and distortion, we either achieve comparable results (least gain score of 0.85) or improve the performance by a margin of $75\%$. We observe the most notable improvements for the calculation time, where we can speed up the computations by a factor of 8 for {\MethodNameBasic} (node2vec) and 7 for {\MethodNameGeneralized} (node2vec). Only for one of the considered cases, the speedup is below $1$, i.e., $0.98$ for {\MethodNameGeneralized}(ctdne) on radoslaw-email dataset. Memory measurements exhibit similar characteristics to other indicators -- {\FrameworkName} performs in most cases no worse than Base methods. In the best case, {\FrameworkName} allows reducing memory consumption up to 4 times. The maximum loss of about $30\%$, compared to vanilla ctdne is encountered with enron-employes, radoslaw-email, and fb-messages, which are small to medium-sized networks.

Results in mean Average Precision exhibit a slightly different nature of the {\FrameworkName} method. For some datasets (bitcoin-otc, fb-messages), we see a drop in the performance. At the same time, distortion in these cases remains at a decent level. We associate such behavior with applying the embedding alignment step, precisely how we choose reference nodes. As explored by us, multi-degree activity function promotes choosing nodes with a similar degree but does not consider their neighborhood. Implementing dataset-specific activity and scoring functions could improve the results. However, one has to take into account the complexity of the function.

\subsubsection{Comparison to Streaming methods}

\begin{figure*}[!ht]
    \centering
    \includegraphics[width=\textwidth]{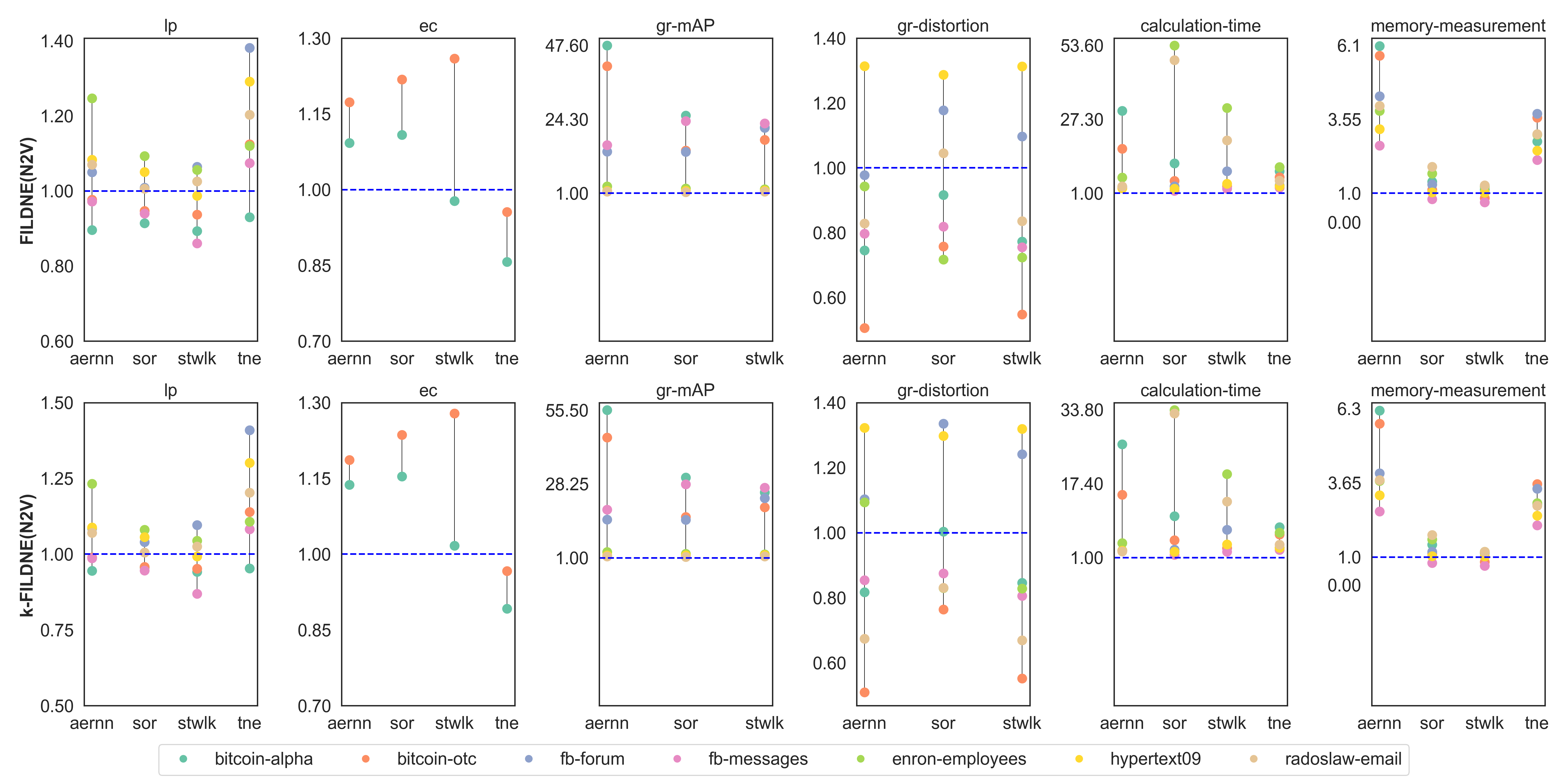}
    \caption{Mean gain scores for {\MethodNameBasic} and {\MethodNameGeneralized} (with node2vec as the Base method) compared to other streaming methods. Points above the dashed line (score equal to $1$) indicate a performance gain, whereas points below indicate loss. Graph reconstruction results are not reported for tNodeEmbed, because the method performs an end-to-end task. We denote tNodeEmbed as \texttt{tne}, DynGraph2vec(AERNN) as \texttt{aernn} and both Online-Node2vec models as \texttt{stwlk} and \texttt{sor}.}
    \label{fig:gain-loss-streaming}
\end{figure*}

We selected Node2vec embeddings as the ones composed by our {\FrameworkName} method because three out of four competitive streaming approaches are built upon random walks. We compute the gain scores in an analogous way to Section \ref{sec:gain-loss-base-methods}. In terms of link prediction, our proposed {\FrameworkName} models achieve comparable results as the other streaming approaches. For edge classification, the tNodeEmbed presents slightly better results than other methods as it is a fine-tuned end-to-end edge classifier. Graph reconstruction measured by mean Average Precision shows the superiority of {\FrameworkName} by a factor of up to $50$. Distortion gain scores vary between $0.55$ and $1.3$, what confirms {\FrameworkName} performance acceptable.
We observe that other streaming methods tend to be slow compared to {\FrameworkName} method, which is reflected on the time gain scores -- embeddings can be computed up to 30 ({\MethodNameGeneralized}) or 50 ({\MethodNameBasic}) times faster. The memory measurements are not surprising -- we provide a comparable utilization to both Online-Node2vec models (StreamWalk and SecondOrder). Note that these were explicitly designed to reduce the memory footprint. At the same time, {\FrameworkName} significantly outperforms other streaming methods.

\subsection{Hyperparameter sensitivity}
In this experiment, we will demonstrate the influence of {\FrameworkName}'s hyperparameters on downstream link prediction tasks in the last snapshot $G_{10}$. We use Node2vec as the Base method in all of the following experiments.

\begin{figure*}[!ht]
    \centering
    \includegraphics[width=\textwidth]{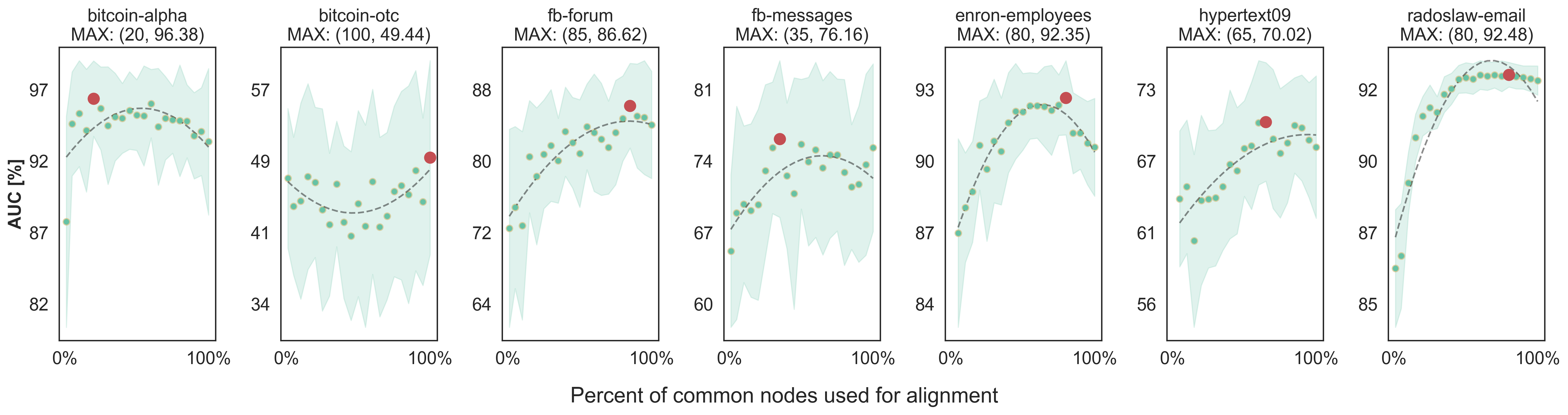}
    \caption{Link prediction task results (AUC [\%]) for different amounts of reference nodes (given as the mean fraction of common nodes between snapshots) The dashed line indicates the trend of these values (fitted using a 2nd order polynomial) and the red dot marks the highest AUC value. Each subtitle presents the name of the dataset and the coordinates of the highest value (percent, AUC).}
    \label{fig:ref-node-percentage}
\end{figure*}

\paragraph{Reference node percentage}
The first hyperparameter we evaluate is the percentage of reference nodes that are used in the alignment step. We proposed in Section \ref{sec:reference_nodes} three selection schemes. Here we focus on the percentage scenario (with values: $1\%, 5\%, 10\%, \ldots, 95\%, 100\%$). We examine if each dataset has an individual percentage that yields the best results in downstream tasks. We plot the mean AUC values and standard deviations in link prediction tasks for each dataset (see Figure \ref{fig:ref-node-percentage}) with {\MethodNameGeneralized} (uniform prior) applied. A single AUC score for a given percentage value is computed as the mean over 30 experiment realizations. Finally, we fit a second-order polynomial (least sq. error) for visualizing the trend on a given dataset and mark in red the point with the highest AUC value.

There is an optimal reference nodes proportion within the exclusive range $(0 \%, 100\%)$ if we observe a concave polynomial approximation. For more complex datasets, e.g., bitcoin-otc, that have a low node and edge Jaccard Index (see table \ref{tab:datasets}) -- representing high dynamics of the network and few common nodes in consecutive windows -- it turns out that the best results are for 100 \% ratio.
    
\begin{figure*}[!ht]
    \centering
    \includegraphics[width=\textwidth]{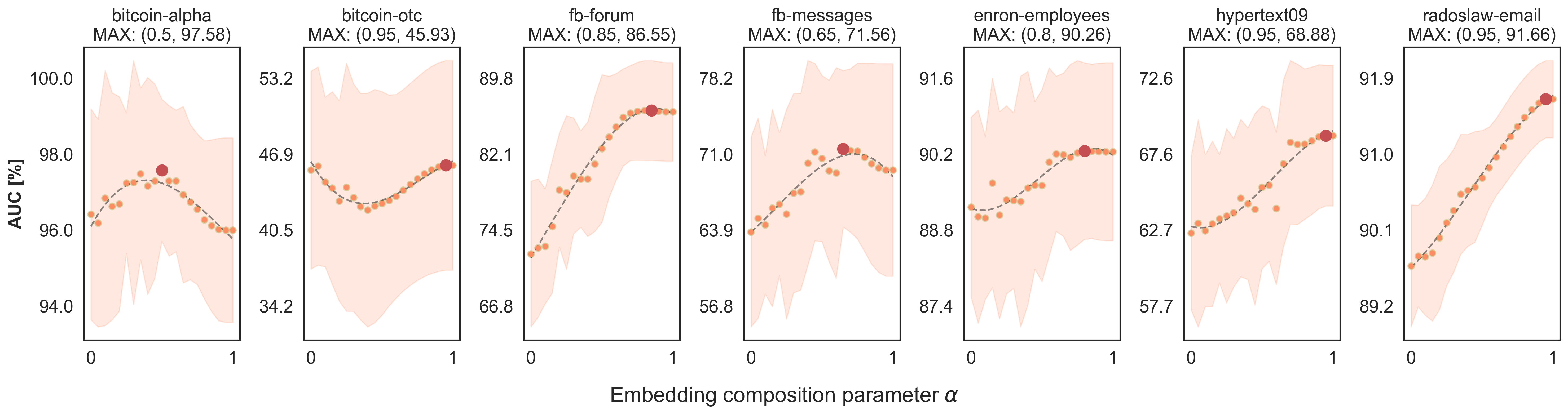}
    \caption{Link prediction task results (AUC [\%]) for different values of $\alpha$. The dashed line indicates the trend of these values (fitted using a 2nd order polynomial) and the red dot marks the highest AUC value. Each subtitle presents the name of the dataset and the coordinates of the highest value ($\alpha$, AUC)}
    \label{fig:basic-icmen-sensitivity}
\end{figure*}

\paragraph{{\MethodNameBasic} alpha parameter}
Next, we examine the impact of  $\alpha$ parameter in {\MethodNameBasic} on AUC, exploring a range from 0 to 1 with a step size of 0.05. The results presented in Figure \ref{fig:basic-icmen-sensitivity} for obtained for 30 experiment repetitions. We utilize the best ratio of reference nodes found in the previous experiment. 

We hypothesized that we should consider both the historical and recent embedding information while constructing node vector representation. Our intuition is confirmed as we can observe different $\alpha$ for distinct datasets. In fb-forum, fb-messages, enron-employees, hypertext09, radoslaw-email, and bitcoin-otc, the past events are critical for performance. We see that greater $\alpha$ values indicate an increase in the embedding quality to a certain level, above which the performance decreases again. For bitcoin-alpha, we have to keep a subtle balance between the present and the past.

\paragraph{{\MethodNameGeneralized} prior}
For {\MethodNameGeneralized}, we change the \textit{prior} parameter and evaluate both possible values (uniform, increase). We noticed that the downstream task performance's change is not significant -- both settings yield similar results. This shows that, after a certain number of observations (internal link prediction), the prior values become irrelevant -- the likelihood computed from the actual data becomes more critical (the property that Maximum Likelihood Estimate is equal to Maximum A Posteriori in the limit).

\subsection{Summary of the results}
We provided an extensive experimental protocol that incorporated various network-related learning tasks, namely link prediction, edge classification, and graph reconstruction. We measured quality in each of those tasks and time/memory consumption on seven benchmark datasets. We can conclude that the experiments' results proved {\FrameworkName} superiority by reducing computation time (up to 50x) and memory consumption (up to 6x), achieving the same quality.
    
\section{Conclusion and future work}
\label{sec:conclusion}
In this paper, we proposed a Framework for Incremental Learning of Dynamic Networks Embedding (FILDNE). It utilizes timestep vectors obtained from any existing node embedding method and produces dynamic representation reducing the computational costs by working on batched data (non-overlapping graph snapshots). We showed experimentally in link prediction, edge classification, and graph reconstruction tasks on seven real-world datasets that FILNDE compared to static, dynamic, and temporal node embedding approaches reduces memory and computational time costs while providing competitive accuracy gains. Moreover, we conducted a hyperparameter sensitivity study and provided insights into how FILDNE's hyperparameters influence the vector representation quality.
In terms of future work, we plan to address FILDNE's limitations, namely: (1) it does not provide a mechanism for online learning (update after the single event) as our method requires batches (graph snapshots); (2) the calibration step requires an overlapping between consecutive snapshots in terms of nodes (there must exist common nodes) -- one can expect a new way to calibrate such snapshots using nodes' structural similarity only; (3) we did not explore how our method works in an attributed environment (i.e., using both attributed graphs and embedding method suitable for such data); (4) in our experimental setup we decided to use snapshots of equal size (in terms of time intervals), but it might be required to extend that scenario.

\section*{Acknowledgments}

The project was partially supported by The National Science Centre, Poland the research projects no. 2016/21/D/ST6/02948 and 2016/21/B/ST6/01463, by the European Union’s Horizon 2020 research and innovation programme under the Marie Skłodowska-Curie grant agreement No 691152 (RENOIR); the Polish Ministry of Science and Higher Education fund for supporting internationally co-financed projects in 2016-2019 (agreement no. 3628/H2020/2016/2) and statutory funds of Department of Computational Intelligence.

\bibliography{related}
\bibliographystyle{splncs04}
\clearpage

\begin{appendices}
    \section{Source code of used methods} 
    We base our expirements on following methods implementations: \\
    \textbf{DynGraph2Vec} - https://github.com/palash1992/DynamicGEM \\
    \textbf{DGI} - https://github.com/PetarV-/DGI \\ 
    \textbf{N2V} - https://github.com/eliorc/node2vec/ \\ 
    \textbf{HOPE} - https://github.com/palash1992/GEM \\ 
    \textbf{LLE} - https://github.com/palash1992/GEM \\
    \textbf{SGE} - https://scikit-learn.org/ \\
    \textbf{OnlineN2V} - https://github.com/ferencberes/online-node2vec \\
    \textbf{TNE} - https://github.com/urielsinger/tNodeEmbed \\
    \textbf{FILDNE} - code of our method will be published upon aceptance   \\

    \section{All methods results (mean)}
    \label{app:full-tables-mean}
        We provide comprehensive experimental results for all examined streaming approaches, Base methods and both {\FrameworkName} variants applied on these Base methods.
        \input{tables_new_fixed/lp/mean-auc-lp}
        \input{tables_new_fixed/ec/mean-auc-ec}
        \input{tables_new_fixed/gr-mAP/mean-gr-mAP}
        \input{tables_new_fixed/gr-distortion/men-gr-distortion}
        \input{tables_new_fixed/calculation-time/mean-calculation-time}
        \input{tables_new_fixed/mm/mean-mm}
\end{appendices}
\end{document}

%% file: tables_new/lp/base-methods-selection-lp.tex
\begin{table*}
    \centering
    \caption{Comparison of \emph{link prediction (AUC [\%])} on cumulative graph embeddings $F_{0, t}$, $t \in \{1, \ldots, 9\}$ across all Base methods. The presented values are the mean AUC over 8 snapshots ($G_{2,3}, \ldots, G_{9, 10}$) and 30 methods' retrains. We select the 3 best methods (LINE, Node2vec, CTDNE) based on the mean ranks and use those methods in further experiments. Underlined values show the highest AUC score for each dataset individually. Methods that did not converged are marked as “$\times$”.}
        \begin{sc}
            \scalebox{0.8}{
\begin{tabular}{lccccccc|c}
\toprule
 &  bitcoin &  bitcoin &  fb &  fb &  enron & hypertext09 & radoslaw & mean rank \\
 & alpha & otc & forum & messages & employees & & email &  \\

\midrule
$\textbf{line}$&  $\underline{66.08}$ &  $\underline{54.69}$ &  $\underline{79.03}$ &  $\underline{70.75}$ &                90.59 &                72.71 &                86.74 &  $\textbf{1.89}$ \\
$\textbf{n2v}$&                65.60 &                52.82 &                75.84 &                65.43 &  $\underline{92.28}$ &  $\underline{73.76}$ &  $\underline{90.15}$ &  $\textbf{1.91}$ \\
$\textbf{ctdne}$&                55.58 &                51.23 &                65.24 &                54.35 &                87.30 &                67.31 &                83.32 &  $\textbf{3.46}$ \\
\hline
hope  &                59.16 &                49.71 &                53.72 &                51.16 &                58.79 &                53.89 &                88.71 &             4.64 \\
dgi   &                55.53 &                50.26 &                54.34 &                53.09 &                59.76 &                57.23 &                70.88 &             4.91 \\
le   &                47.46 &                49.73 &                54.17 &                51.81 &                58.23 &                55.01 &                77.99 &             5.07 \\
lle   &                    $\times$ &                      $\times$  &                55.69 &                      $\times$ &                59.96 &                58.13 &                54.87 &             5.44 \\
\bottomrule
\end{tabular}
            }
        \end{sc}

\label{tab:base-method-selection}
\end{table*}

%% file: tables_new/lp/mean-auc-lp.tex
\begin{table*}
    \centering
    \caption{Comparison of {\FrameworkName} and other methods in \emph{link prediction task (AUC [\%])}. The presented values are the mean AUC scores over 8 graph snapshots and 30 methods' retrains. We also report the mean ranks of all methods in this experiment. Methods marked in bold are the 3 best methods based on the mean rank. Underlined values show the highest AUC score for each dataset individually.}
        \begin{sc}
            \scalebox{0.75}{
\begin{tabular}{lccccccc|c}
\toprule
 &  bitcoin &  bitcoin &  fb &  fb &  enron & hypertext09 & radoslaw & mean rank \\
 & alpha & otc & forum & messages & employees & & email &  \\
\midrule
$\textbf{n2v}$&                65.60 &                52.82 &                75.84 &                65.43 &                92.28 &                73.76 &                90.15 &  $\textbf{5.43}$ \\
line                              &                66.08 &                54.69 &  $\underline{79.03}$ &                70.75 &                90.59 &                72.71 &                86.74 &             5.77 \\
ctdne                             &                55.58 &                51.23 &                65.24 &                54.35 &                87.30 &                67.31 &                83.32 &             9.61 \\
\midrule
{\MethodNameBasicLC}(n2v)         &                59.25 &                53.36 &                73.94 &                64.27 &  $\underline{92.43}$ &                71.35 &                91.67 &             5.75 \\
$\textbf{{\MethodNameBasicLC}(line)}$&                58.28 &  $\underline{62.39}$ &                74.11 &  $\underline{76.11}$ &                92.00 &  $\underline{73.87}$ &                93.42 &  $\textbf{4.21}$ \\
{\MethodNameBasicLC}(ctdne)       &                55.39 &                55.97 &                66.10 &                55.86 &                88.33 &                65.51 &                85.64 &             9.30 \\
\midrule
{\MethodNameGeneralizedLC}(n2v)   &                62.22 &                53.93 &                75.95 &                65.09 &                91.43 &                71.71 &                91.66 &             5.55 \\
$\textbf{{\MethodNameGeneralizedLC}(line)}$&                59.41 &                59.66 &                75.06 &                74.83 &                91.92 &                73.27 &  $\underline{93.62}$ &  $\textbf{4.48}$ \\
{\MethodNameGeneralizedLC}(ctdne) &                53.52 &                53.38 &                61.24 &                57.01 &                87.85 &                64.44 &                85.54 &            10.00 \\
\midrule
dynGraph2Vec(aernn)               &                67.88 &                58.67 &                70.50 &                66.93 &                74.66 &                66.40 &                85.72 &             8.38 \\
tNodeEmbed                        &                68.68 &                48.40 &                55.96 &                60.97 &                83.57 &                55.79 &                76.87 &             9.61 \\
online-n2v(streamwalk)            &  $\underline{69.59}$ &                59.66 &                73.21 &                69.55 &                84.84 &                68.36 &                91.13 &             6.68 \\
online-n2v(secondorder)           &                68.70 &                58.33 &                69.34 &                75.31 &                87.59 &                72.44 &                89.41 &             6.23 \\
\bottomrule
\end{tabular}
            }
        \end{sc}
    
\label{tab:mean-auc-lp}
\end{table*}

%% file: tables_new/ec/mean-auc-ec.tex
\begin{table}[H]
    \centering
    \caption{Comparison of {\FrameworkName} and other methods in \emph{edge classification task (AUC [\%])}. The presented values are the mean AUC scores over 8 graph snapshots and 30 methods' retrains. We also report the mean ranks of all methods in this experiment. Methods marked in bold are the 3 best methods based on the mean rank. Underlined values show the highest AUC score for each dataset individually.}
        \begin{sc}
            \scalebox{0.75}{
\begin{tabular}{lcc|c}
\toprule
&  bitcoin &  bitcoin & mean rank \\
& alpha & otc & \\
\midrule
n2v                               &                65.11 &                60.92 &             7.25 \\
line                              &                63.09 &                64.39 &             6.44 \\
ctdne                             &                63.81 &                55.92 &             8.56 \\
\midrule
{\MethodNameBasicLC}(n2v)         &                61.57 &                65.47 &             6.56 \\
$\textbf{{\MethodNameBasicLC}(line)}$&                63.91 &                66.56 &  $\textbf{5.00}$ \\
{\MethodNameBasicLC}(ctdne)       &                59.28 &                57.85 &             8.56 \\
\midrule
{\MethodNameGeneralizedLC}(n2v)   &                64.10 &                66.23 &             5.25 \\
$\textbf{{\MethodNameGeneralizedLC}(line)}$&                63.92 &                66.57 &  $\textbf{4.94}$ \\
{\MethodNameGeneralizedLC}(ctdne) &                58.25 &                59.84 &             8.06 \\
\midrule
dynGraph2Vec(aernn)               &                56.40 &                56.14 &            10.12 \\
$\textbf{tNodeEmbed}$&  $\underline{71.88}$ &  $\underline{69.32}$ &  $\textbf{3.12}$ \\
online-n2v(streamwalk)            &                56.73 &                56.70 &             9.25 \\
online-n2v(secondorder)           &                63.39 &                53.93 &             7.88 \\
\bottomrule
\end{tabular}

            }
        \end{sc}
    
\label{tab:mean-auc-ec}
\end{table}

%% file: tables_new/gr-mAP/mean-gr-mAP.tex
\begin{table*}
    \centering
    \caption{Comparison of {\FrameworkName} and other methods in \emph{graph reconstruction task (mAP [\%])}. The presented values are the mean mAP scores over 8 graph snapshots and 30 methods' retrains. We also report the mean ranks of all methods in this experiment. Methods marked in bold are the 3 best methods based on the mean rank. Underlined values show the highest mAP score for each dataset individually.}
        \begin{sc}
            \scalebox{0.75}{
\begin{tabular}{lccccccc|c}
\toprule
 &  bitcoin &  bitcoin &  fb &  fb &  enron & hypertext09 & radoslaw & mean rank \\
 & alpha & otc & forum & messages & employees & & email &  \\
\midrule
$\textbf{n2v}$&  $\underline{65.99}$ &  $\underline{70.18}$ &                23.16 &  $\underline{67.44}$ &  $\underline{70.46}$ &  $\underline{55.95}$ &  $\underline{42.12}$ &  $\textbf{1.36}$ \\
line                              &                10.83 &                10.81 &                 6.42 &                 9.90 &                45.35 &                41.90 &                39.75 &             7.52 \\
ctdne                             &                35.65 &                47.24 &                 7.85 &                22.84 &                55.15 &                54.80 &                31.42 &             4.61 \\
\midrule
$\textbf{{\MethodNameBasicLC}(n2v)}$&                28.73 &                26.40 &                30.98 &                30.00 &                63.90 &                49.45 &                40.55 &  $\textbf{3.55}$ \\
{\MethodNameBasicLC}(line)        &                14.70 &                18.02 &                12.16 &                 7.12 &                41.71 &                38.24 &                38.38 &             7.61 \\
{\MethodNameBasicLC}(ctdne)       &                18.20 &                24.27 &                 9.53 &                11.59 &                49.39 &                45.92 &                36.84 &             6.30 \\
\midrule
$\textbf{{\MethodNameGeneralizedLC}(n2v)}$&                33.46 &                29.30 &  $\underline{33.17}$ &                34.66 &                64.21 &                53.05 &                40.92 &  $\textbf{2.41}$ \\
{\MethodNameGeneralizedLC}(line)  &                15.39 &                25.02 &                15.58 &                 8.92 &                44.43 &                38.82 &                38.08 &             6.66 \\
{\MethodNameGeneralizedLC}(ctdne) &                16.85 &                27.46 &                13.93 &                20.39 &                55.10 &                48.48 &                37.26 &             4.98 \\
\midrule
dynGraph2Vec(aernn)               &                 0.60 &                 0.66 &                 2.22 &                 1.84 &                20.46 &                27.75 &                26.51 &            11.32 \\
online-n2v(streamwalk)            &                 1.08 &                 1.83 &                 2.22 &                 1.25 &                25.85 &                28.03 &                29.60 &            10.70 \\
online-n2v(secondorder)           &                 1.38 &                 1.54 &                 1.46 &                 1.28 &                29.09 &                27.88 &                27.11 &            10.98 \\
\bottomrule
\end{tabular}
            }
        \end{sc}
    
\label{tab:mean-gr-map}
\end{table*}

%% file: tables_new/gr-distortion/mean-gr-distortion.tex
\begin{table*}
    \centering
    \caption{Comparison of {\FrameworkName} and other methods in \emph{graph reconstruction task (distortion)}. The presented values are the mean distortion scores over 8 graph snapshots and 30 methods' retrains. We also report the mean ranks of all methods in this experiment. Methods marked in bold are the 3 best methods based on the mean rank. Underlined values show the lowest (best) distortion score for each dataset individually.}
        \begin{sc}
            \scalebox{0.75}{
\begin{tabular}{lccccccc|c}
\toprule
 &  bitcoin &  bitcoin &  fb &  fb &  enron & hypertext09 & radoslaw & mean rank \\
 & alpha & otc & forum & messages & employees & & email &  \\
\midrule
n2v                               &                0.81 &                1.29 &                0.61 &                0.93 &                0.66 &                0.38 &                0.57 &             8.41 \\
line                              &                0.87 &                0.94 &                0.50 &                0.59 &                0.65 &                0.33 &                0.80 &             7.52 \\
ctdne                             &                0.93 &                1.70 &                0.76 &                0.89 &                0.81 &                0.36 &                0.65 &            10.12 \\
\midrule
{\MethodNameBasicLC}(n2v)         &                0.70 &                1.05 &                0.47 &                0.74 &                0.69 &                0.30 &                0.54 &             6.21 \\
{\MethodNameBasicLC}(line)        &                0.65 &                0.65 &                0.41 &                0.54 &                0.67 &                0.31 &                0.70 &             5.43 \\
{\MethodNameBasicLC}(ctdne)       &                0.85 &                1.70 &                0.65 &                0.93 &                0.84 &                0.32 &                0.66 &             9.45 \\
\midrule
{\MethodNameGeneralizedLC}(n2v)   &                0.64 &                1.04 &                0.44 &                0.72 &                0.63 &                0.30 &                0.67 &             5.59 \\
$\textbf{{\MethodNameGeneralizedLC}(line)}$&                0.65 &                0.73 &  $\underline{0.33}$ &                0.50 &                0.66 &                0.31 &                0.74 &  $\textbf{5.05}$ \\
{\MethodNameGeneralizedLC}(ctdne) &                0.83 &                1.65 &                0.46 &                0.75 &                0.68 &  $\underline{0.30}$ &                0.66 &             6.84 \\
\midrule
$\textbf{dynGraph2Vec(aernn)}$&  $\underline{0.52}$ &  $\underline{0.53}$ &                0.44 &                0.50 &                0.62 &                0.40 &  $\underline{0.45}$ &  $\textbf{4.20}$ \\
online-n2v(streamwalk)            &                0.64 &                0.80 &                0.52 &                0.53 &  $\underline{0.47}$ &                0.39 &                0.56 &             5.39 \\
$\textbf{online-n2v(secondorder)}$&                0.54 &                0.57 &                0.48 &  $\underline{0.49}$ &                0.49 &                0.40 &                0.45 &  $\textbf{3.79}$ \\
\bottomrule
\end{tabular}
            }
        \end{sc}
    
\label{tab:mean-gr-distortion}
\end{table*}

%% file: tables_new/calculation-time/mean-calculation-time.tex
\begin{table*}
    \centering
    \caption{Comparison of {\FrameworkName} and other methods in \emph{embeddings calculation time [s]}. The presented values are the mean calculation time over 8 graph snapshots and 30 methods' retrains. We also report the mean ranks of all methods in this experiment. Methods marked in bold are the 3 best methods based on the mean rank. Underlined values show the lowest calculation time for each dataset individually.}
        \begin{sc}
            \scalebox{0.75}{
\begin{tabular}{lccccccc|c}
\toprule
 &  bitcoin &  bitcoin &  fb &  fb &  enron & hypertext09 & radoslaw & mean rank \\
 & alpha & otc & forum & messages & employees & & email &  \\
\midrule
n2v                               &               46.71 &               133.83 &               24.94 &                75.72 &                4.40 &                9.52 &               11.90 &             7.20 \\
line                              &               70.78 &                63.10 &              103.97 &               114.16 &              114.84 &              105.17 &               58.56 &            11.79 \\
ctdne                             &               16.25 &                61.41 &                3.98 &                77.44 &               16.82 &                4.39 &                2.65 &             5.27 \\
\midrule
$\textbf{{\MethodNameBasicLC}(n2v)}$&                7.87 &                22.26 &                8.44 &  $\underline{27.62}$ &  $\underline{2.35}$ &                4.79 &                4.65 &  $\textbf{2.75}$ \\
{\MethodNameBasicLC}(line)        &               42.34 &                41.66 &               80.53 &                82.18 &               84.43 &               83.29 &               43.50 &             9.07 \\
$\textbf{{\MethodNameBasicLC}(ctdne)}$&  $\underline{7.84}$ &  $\underline{18.11}$ &  $\underline{1.91}$ &                33.67 &                8.69 &  $\underline{2.63}$ &  $\underline{0.68}$ &  $\textbf{1.98}$ \\
\midrule
{\MethodNameGeneralizedLC}(n2v)   &                8.61 &                24.04 &                9.90 &                30.69 &                3.71 &                5.31 &                6.40 &             3.93 \\
{\MethodNameGeneralizedLC}(line)  &               43.10 &                43.20 &               81.74 &                85.02 &               85.67 &               83.81 &               45.09 &            10.11 \\
$\textbf{{\MethodNameGeneralizedLC}(ctdne)}$&                8.59 &                19.88 &                3.33 &                36.37 &                9.96 &                3.15 &                2.71 &  $\textbf{3.20}$ \\
\midrule
dynGraph2Vec(aernn)               &              144.71 &               341.14 &               25.54 &                84.10 &               15.89 &               13.35 &               15.47 &             7.82 \\
tNodeEmbed                        &               52.46 &               143.07 &               31.32 &                84.88 &               24.96 &               16.90 &               24.85 &             9.48 \\
online-n2v(streamwalk)            &               67.76 &               112.97 &               27.44 &                52.09 &              130.27 &               12.50 &              212.59 &             9.43 \\
online-n2v(secondorder)           &               20.78 &                72.98 &               70.51 &                74.15 &               76.83 &               20.92 &               86.89 &             8.98 \\
\bottomrule
\end{tabular}

            }
        \end{sc}
    
\label{tab:mean-calculation-time}
\end{table*}

%% file: tables_new/mm/mean-mm.tex
\begin{table*}
    \centering
    \caption{Comparison of {\FrameworkName} and other methods in \emph{max. memory utilization [MB]}. The presented values are the mean max. memory utilization during embeddings calculation over 8 graph snapshots and 30 methods' retrains. We also report the mean ranks of all methods in this experiment. Methods marked in bold are the 3 best methods based on the mean rank. Underlined values show the lowest memory utilization for each dataset individually.}
        \begin{sc}
            \scalebox{0.75}{
\begin{tabular}{lccccccc|c}
\toprule
 &  bitcoin &  bitcoin &  fb &  fb &  enron & hypertext09 & radoslaw & mean rank \\
 & alpha & otc & forum & messages & employees & & email &  \\
\midrule
n2v                               &                580.09 &                317.12 &                552.29 &                360.35 &                212.27 &  $\underline{218.45}$ &                227.75 &             3.71 \\
line                              &                917.17 &                947.08 &               1291.09 &               1309.96 &               1286.30 &               1277.53 &                920.00 &            12.16 \\
ctdne                             &               1751.63 &               1262.22 &               1327.77 &               1113.66 &                344.43 &                320.13 &                361.23 &             9.07 \\
\midrule
$\textbf{{\MethodNameBasicLC}(n2v)}$&                233.96 &                344.62 &  $\underline{207.13}$ &                385.37 &  $\underline{206.60}$ &                241.49 &  $\underline{203.49}$ &  $\textbf{2.23}$ \\
{\MethodNameBasicLC}(line)        &                888.59 &                899.76 &               1275.47 &               1279.96 &               1271.36 &               1269.56 &                889.90 &            10.45 \\
{\MethodNameBasicLC}(ctdne)       &                436.55 &                849.80 &                338.78 &               1270.43 &                514.53 &                384.26 &                300.85 &             6.80 \\
\midrule
$\textbf{{\MethodNameGeneralizedLC}(n2v)}$&  $\underline{224.82}$ &                344.28 &                225.41 &                387.52 &                213.02 &                241.50 &                218.23 &  $\textbf{2.62}$ \\
{\MethodNameGeneralizedLC}(line)  &                888.82 &                901.10 &               1276.48 &               1281.39 &               1271.71 &               1269.49 &                889.21 &            10.66 \\
{\MethodNameGeneralizedLC}(ctdne) &                437.24 &                848.13 &                340.99 &               1272.02 &                514.99 &                385.14 &                300.77 &             7.09 \\
\midrule
dynGraph2Vec(aernn)               &               1373.12 &               1934.64 &                899.86 &               1023.55 &                793.38 &                774.93 &                816.70 &            10.20 \\
tNodeEmbed                        &                643.19 &               1225.00 &                774.82 &                827.67 &                624.12 &                598.00 &                618.93 &             8.54 \\
online-n2v(streamwalk)            &                321.29 &                381.68 &                269.47 &                307.71 &                347.81 &                249.34 &                390.14 &             4.70 \\
$\textbf{online-n2v(secondorder)}$&                260.86 &  $\underline{284.38}$ &                258.63 &  $\underline{265.96}$ &                253.60 &                246.26 &                259.93 &  $\textbf{2.77}$ \\
\bottomrule
\end{tabular}
            }
        \end{sc}
    
\label{tab:mean-mm}
\end{table*}

%% file: tables_new_fixed/lp/mean-auc-lp.tex
\begin{table*}
    \centering
    \caption{Comparison of {\FrameworkName} and other methods in \textbf{link prediction task (AUC)}. The presented values are the mean AUC scores over 8 graph snapshots and 30 methods' retrains. We also report the mean ranks of all methods in this experiment. Methods marked in bold are the 3 best methods based on the mean rank. Underlined values show the highest AUC score for each dataset individually.}
        \begin{sc}
            \scalebox{0.75}{
\begin{tabular}{lccccccc|c}
\toprule
&  bitcoin &  bitcoin &  fb &  fb &  enron & hypertext09 & radoslaw & mean rank  \\
& alpha & otc & forum & messages & employees & & email & \\
\midrule
$\textbf{n2v}$&                65.60 &                52.82 &                75.84 &                65.43 &                92.28 &                73.76 &                90.15 &  $\textbf{6.29}$ \\
line                              &                66.08 &                54.69 &  $\underline{79.03}$ &                70.75 &                90.59 &                72.71 &                86.74 &             6.48 \\
ctdne                             &                55.58 &                51.23 &                65.24 &                54.35 &                87.30 &                67.31 &                83.32 &            11.79 \\
hope                              &                59.16 &                49.71 &                53.72 &                51.16 &                58.79 &                53.89 &                88.71 &            16.32 \\
dgi                               &                55.53 &                50.26 &                54.34 &                53.09 &                59.76 &                57.23 &                70.88 &            17.36 \\
le                                &                47.46 &                49.73 &                54.17 &                51.81 &                58.23 &                55.01 &                77.99 &            18.02 \\
lle                               &             $\times$ &             $\times$ &                55.69 &             $\times$ &                59.96 &                58.13 &                54.87 &            19.69 \\
\midrule
{\MethodNameBasicLC}(n2v)         &                59.25 &                53.36 &                73.94 &                64.27 &  $\underline{92.43}$ &                71.35 &                91.67 &             6.98 \\
$\textbf{{\MethodNameBasicLC}(line)}$&                58.28 &  $\underline{62.39}$ &                74.11 &  $\underline{76.11}$ &                92.00 &  $\underline{73.87}$ &                93.42 &  $\textbf{4.79}$ \\
{\MethodNameBasicLC}(ctdne)       &                55.39 &                55.97 &                66.10 &                55.86 &                88.33 &                65.51 &                85.64 &            11.00 \\
{\MethodNameBasicLC}(hope)        &                49.83 &                50.21 &                53.57 &                55.20 &                59.08 &                59.37 &                88.16 &            15.57 \\
{\MethodNameBasicLC}(dgi)         &                53.98 &                51.48 &                55.57 &                53.32 &                63.56 &                59.89 &                68.78 &            15.77 \\
{\MethodNameBasicLC}(le)         &                49.39 &                50.59 &                55.27 &                49.15 &                62.12 &                58.03 &                80.72 &            16.79 \\
{\MethodNameBasicLC}(lle)         &             $\times$ &             $\times$ &                56.19 &             $\times$ &                62.74 &                58.99 &                55.53 &            18.38 \\
\midrule
{\MethodNameGeneralizedLC}(n2v)   &                62.22 &                53.93 &                75.95 &                65.09 &                91.43 &                71.71 &                91.66 &             6.52 \\
$\textbf{{\MethodNameGeneralizedLC}(line)}$&                59.41 &                59.66 &                75.06 &                74.83 &                91.92 &                73.27 &  $\underline{93.62}$ &  $\textbf{5.21}$ \\
{\MethodNameGeneralizedLC}(ctdne) &                53.52 &                53.38 &                61.24 &                57.01 &                87.85 &                64.44 &                85.54 &            12.25 \\
{\MethodNameGeneralizedLC}(hope)  &                53.60 &                49.50 &                53.35 &                54.08 &                53.30 &                58.36 &                78.47 &            17.39 \\
{\MethodNameGeneralizedLC}(dgi)   &                51.32 &                47.97 &                55.71 &                53.86 &                61.75 &                58.83 &                66.27 &            17.23 \\
{\MethodNameGeneralizedLC}(le)   &                51.60 &                50.59 &                54.12 &                49.41 &                61.40 &                57.98 &                60.27 &            17.93 \\
{\MethodNameGeneralizedLC}(lle)   &             $\times$ &             $\times$ &                55.84 &             $\times$ &                61.33 &                57.73 &                56.53 &            19.44 \\
\midrule
dynGraph2Vec(aernn)               &                67.88 &                58.67 &                70.50 &                66.93 &                74.66 &                66.40 &                85.72 &             9.41 \\
tNodeEmbed                        &                68.68 &                48.40 &                55.96 &                60.97 &                83.57 &                55.79 &                76.87 &            13.93 \\
online-n2v(streamwalk)            &  $\underline{69.59}$ &                59.66 &                73.21 &                69.55 &                84.84 &                68.36 &                91.13 &             7.50 \\
online-n2v(secondorder)           &                68.70 &                58.33 &                69.34 &                75.31 &                87.59 &                72.44 &                89.41 &             6.77 \\
\bottomrule
\end{tabular}
            }
        \end{sc}
    
\label{table:mean-auc-lp-all}
\end{table*}

%% file: tables_new_fixed/ec/mean-auc-ec.tex
\begin{table*}
    \centering
    \caption{Comparison of {\FrameworkName} and other methods in \textbf{edge classification task (AUC)}. The presented values are the mean AUC scores over 8 graph snapshots and 30 methods' retrains. We also report the mean ranks of all methods in this experiment. Methods marked in bold are the 3 best methods based on the mean rank. Underlined values show the highest AUC score for each dataset individually.}
        \begin{sc}
            \scalebox{0.75}{
\begin{tabular}{lcc|r}
\toprule
&  bitcoin &  bitcoin & mean rank \\
& alpha & otc & \\
\midrule
n2v                               &                65.11 &                60.92 &             8.69 \\
line                              &                63.09 &                64.39 &             7.81 \\
ctdne                             &                63.81 &                55.92 &            10.56 \\
hope                              &                57.76 &                43.04 &            17.25 \\
dgi                               &                57.58 &                59.39 &            12.00 \\
le                                &                53.60 &                44.60 &            16.41 \\
lle                               &             $\times$ &             $\times$ &         $\times$ \\
\midrule
{\MethodNameBasicLC}(n2v)         &                61.57 &                65.47 &             7.75 \\
$\textbf{{\MethodNameBasicLC}(line)}$&                63.91 &                66.56 &  $\textbf{6.19}$ \\
{\MethodNameBasicLC}(ctdne)       &                59.28 &                57.85 &            10.94 \\
{\MethodNameBasicLC}(hope)        &                58.89 &                50.77 &            15.03 \\
{\MethodNameBasicLC}(dgi)         &                58.26 &                52.81 &            14.38 \\
{\MethodNameBasicLC}(le)         &                52.31 &                54.63 &            16.25 \\
{\MethodNameBasicLC}(lle)         &             $\times$ &             $\times$ &         $\times$ \\
\midrule
{\MethodNameGeneralizedLC}(n2v)   &                64.10 &                66.23 &             6.25 \\
$\textbf{{\MethodNameGeneralizedLC}(line)}$&                63.92 &                66.57 &  $\textbf{6.12}$ \\
{\MethodNameGeneralizedLC}(ctdne) &                58.25 &                59.84 &            10.19 \\
{\MethodNameGeneralizedLC}(hope)  &                57.87 &                49.84 &            15.00 \\
{\MethodNameGeneralizedLC}(dgi)   &                58.29 &                54.61 &            12.94 \\
{\MethodNameGeneralizedLC}(le)   &                47.80 &                49.87 &            18.50 \\
{\MethodNameGeneralizedLC}(lle)   &             $\times$ &             $\times$ &         $\times$ \\
\midrule
dynGraph2Vec(aernn)               &                56.40 &                56.14 &            13.62 \\
$\textbf{tNodeEmbed}$&  $\underline{71.88}$ &  $\underline{69.32}$ &  $\textbf{3.62}$ \\
online-n2v(streamwalk)            &                56.73 &                56.70 &            12.81 \\
online-n2v(secondorder)           &                63.39 &                53.93 &            10.69 \\
\bottomrule
\end{tabular}

            }
        \end{sc}
    
\label{table:mean-auc-ec-all}
\end{table*}

%% file: tables_new_fixed/gr-mAP/mean-gr-mAP.tex
\begin{table*}
    \centering
    \caption{Comparison of {\FrameworkName} and other methods in \textbf{graph reconstruction task (mAP)}. The presented values are the mean mAP scores over 8 graph snapshots and 30 methods' retrains. We also report the mean ranks of all methods in this experiment. Methods marked in bold are the 3 best methods based on the mean rank. Underlined values show the highest mAP score for each dataset individually.}
        \begin{sc}
            \scalebox{0.75}{
\begin{tabular}{lccccccc|c}
\toprule
&  bitcoin &  bitcoin &  fb &  fb &  enron & hypertext09 & radoslaw &  mean rank \\
& alpha & otc & forum & messages & employees & & email &\\
\midrule
$\textbf{n2v}$&  $\underline{65.99}$ &  $\underline{70.18}$ &                23.16 &  $\underline{67.44}$ &  $\underline{70.46}$ &  $\underline{55.95}$ &                42.12 &  $\textbf{1.84}$ \\
line                              &                10.83 &                10.81 &                 6.42 &                 9.90 &                45.35 &                41.90 &                39.75 &             8.05 \\
ctdne                             &                35.65 &                47.24 &                 7.85 &                22.84 &                55.15 &                54.80 &                31.42 &             5.25 \\
hope                              &                 0.51 &                 0.64 &                 2.43 &                 1.39 &                13.20 &                28.10 &                27.06 &            18.32 \\
dgi                               &                 0.66 &                 0.66 &                 2.55 &                 1.45 &                12.44 &                27.11 &                26.91 &            18.48 \\
le                                &                 0.81 &                 1.12 &                 2.71 &                 1.22 &                11.89 &                28.04 &  $\underline{76.53}$ &            14.00 \\
lle                               &             $\times$ &             $\times$ &                 2.65 &             $\times$ &                11.98 &                27.16 &                48.62 &            15.38 \\
\midrule
$\textbf{{\MethodNameBasicLC}(n2v)}$&                28.73 &                26.40 &                30.98 &                30.00 &                63.90 &                49.45 &                40.55 &  $\textbf{4.04}$ \\
{\MethodNameBasicLC}(line)        &                14.70 &                18.02 &                12.16 &                 7.12 &                41.71 &                38.24 &                38.38 &             8.18 \\
{\MethodNameBasicLC}(ctdne)       &                18.20 &                24.27 &                 9.53 &                11.59 &                49.39 &                45.92 &                36.84 &             6.86 \\
{\MethodNameBasicLC}(hope)        &                 0.59 &                 0.67 &                 2.50 &                 1.46 &                12.90 &                27.42 &                27.55 &            18.27 \\
{\MethodNameBasicLC}(dgi)         &                 0.77 &                 0.80 &                 2.88 &                 1.68 &                12.69 &                27.20 &                29.66 &            15.50 \\
{\MethodNameBasicLC}(le)         &                 0.76 &                 0.94 &                 2.71 &                 1.29 &                13.08 &                27.64 &                60.00 &            14.09 \\
{\MethodNameBasicLC}(lle)         &             $\times$ &             $\times$ &                 2.58 &             $\times$ &                11.21 &                28.12 &                26.34 &            19.34 \\
\midrule
$\textbf{{\MethodNameGeneralizedLC}(n2v)}$&                33.46 &                29.30 &  $\underline{33.17}$ &                34.66 &                64.21 &                53.05 &                40.92 &  $\textbf{2.89}$ \\
{\MethodNameGeneralizedLC}(line)  &                15.39 &                25.02 &                15.58 &                 8.92 &                44.43 &                38.82 &                38.08 &             7.23 \\
{\MethodNameGeneralizedLC}(ctdne) &                16.85 &                27.46 &                13.93 &                20.39 &                55.10 &                48.48 &                37.26 &             5.52 \\
{\MethodNameGeneralizedLC}(hope)  &                 0.59 &                 0.70 &                 2.42 &                 1.48 &                13.28 &                29.80 &                23.92 &            17.43 \\
{\MethodNameGeneralizedLC}(dgi)   &                 0.93 &                 0.94 &                 3.24 &                 2.06 &                12.72 &                27.76 &                29.04 &            14.04 \\
{\MethodNameGeneralizedLC}(le)   &                 0.73 &                 0.90 &                 2.81 &                 1.40 &                12.84 &                29.05 &                41.80 &            13.79 \\
{\MethodNameGeneralizedLC}(lle)   &             $\times$ &             $\times$ &                 2.98 &             $\times$ &                12.58 &                28.63 &                31.78 &            14.91 \\
\midrule
dynGraph2Vec(aernn)               &                 0.60 &                 0.66 &                 2.22 &                 1.84 &                20.46 &                27.75 &                26.51 &            17.07 \\
online-n2v(streamwalk)            &                 1.08 &                 1.83 &                 2.22 &                 1.25 &                25.85 &                28.03 &                29.60 &            15.20 \\
online-n2v(secondorder)           &                 1.38 &                 1.54 &                 1.46 &                 1.28 &                29.09 &                27.88 &                27.11 &            16.04 \\
\bottomrule
\end{tabular}

            }
        \end{sc}
    
\label{table:mean-gr-map-all}
\end{table*}

%% file: tables_new_fixed/gr-distortion/men-gr-distortion.tex
\begin{table*}
    \centering
    \caption{Comparison of {\FrameworkName} and other methods in graph reconstruction task (distortion). The presented values are the mean distortion scores over 8 graph snapshots and 30 methods' retrains. We also report the mean ranks of all methods in this experiment. Methods marked in bold are the 3 best methods based on the mean rank. Underlined values show the highest distortion score for each dataset individually.}
        \begin{sc}
            \scalebox{0.75}{
\begin{tabular}{lccccccc|c}
\toprule
&  bitcoin &  bitcoin &  fb &  fb &  enron & hypertext09 & radoslaw  & mean rank \\
& alpha & otc & forum & messages & employees & & email \\
\midrule
n2v                               &                0.81 &                1.29 &                0.61 &                0.93 &                0.66 &                0.38 &                0.57 &            17.66 \\
line                              &                0.87 &                0.94 &                0.50 &                0.59 &                0.65 &                0.33 &                0.80 &            15.93 \\
ctdne                             &                0.93 &                1.70 &                0.76 &                0.89 &                0.81 &                0.36 &                0.65 &            19.41 \\
hope                              &                0.58 &                0.60 &                0.36 &                0.48 &                0.48 &                0.34 &                0.44 &             8.93 \\
dgi                               &  $\underline{0.51}$ &                0.55 &                0.36 &                0.61 &                0.49 &                0.34 &  $\underline{0.38}$ &             7.38 \\
le                                &                0.58 &                0.58 &                0.33 &                0.45 &                0.47 &                0.33 &                0.48 &             8.43 \\
lle                               &            $\times$ &            $\times$ &                0.54 &            $\times$ &                0.77 &                0.35 &                0.77 &            18.91 \\
\midrule
{\MethodNameBasicLC}(n2v)         &                0.70 &                1.05 &                0.47 &                0.74 &                0.69 &                0.30 &                0.54 &            14.29 \\
{\MethodNameBasicLC}(line)        &                0.65 &                0.65 &                0.41 &                0.54 &                0.67 &                0.31 &                0.70 &            12.70 \\
{\MethodNameBasicLC}(ctdne)       &                0.85 &                1.70 &                0.65 &                0.93 &                0.84 &                0.32 &                0.66 &            18.54 \\
{\MethodNameBasicLC}(hope)        &                0.56 &                0.60 &                0.40 &                0.48 &                0.48 &                0.34 &                0.42 &             8.89 \\
{\MethodNameBasicLC}(dgi)         &                0.53 &                0.72 &                0.30 &                0.55 &                0.47 &                0.33 &                0.38 &             7.68 \\
$\textbf{{\MethodNameBasicLC}(le)}$&                0.55 &                0.65 &                0.28 &  $\underline{0.38}$ &  $\underline{0.41}$ &                0.34 &                0.47 &  $\textbf{6.96}$ \\
{\MethodNameBasicLC}(lle)         &            $\times$ &            $\times$ &                0.46 &            $\times$ &                0.88 &                0.35 &                0.81 &            19.00 \\
\midrule
{\MethodNameGeneralizedLC}(n2v)   &                0.64 &                1.04 &                0.44 &                0.72 &                0.63 &                0.30 &                0.67 &            12.91 \\
{\MethodNameGeneralizedLC}(line)  &                0.65 &                0.73 &                0.33 &                0.50 &                0.66 &                0.31 &                0.74 &            11.61 \\
{\MethodNameGeneralizedLC}(ctdne) &                0.83 &                1.65 &                0.46 &                0.75 &                0.68 &  $\underline{0.30}$ &                0.66 &            14.73 \\
{\MethodNameGeneralizedLC}(hope)  &                0.57 &                0.60 &                0.43 &                0.48 &                0.48 &                0.32 &                0.40 &             8.50 \\
$\textbf{{\MethodNameGeneralizedLC}(dgi)}$&                0.52 &                0.72 &                0.30 &                0.51 &                0.49 &                0.33 &                0.42 &  $\textbf{7.30}$ \\
$\textbf{{\MethodNameGeneralizedLC}(le)}$&                0.55 &                0.61 &  $\underline{0.28}$ &                0.38 &                0.43 &                0.32 &                0.43 &  $\textbf{5.52}$ \\
{\MethodNameGeneralizedLC}(lle)   &            $\times$ &            $\times$ &                0.40 &            $\times$ &                0.60 &                0.33 &                0.60 &            11.78 \\
\midrule
dynGraph2Vec(aernn)               &                0.52 &  $\underline{0.53}$ &                0.44 &                0.50 &                0.62 &                0.40 &                0.45 &            10.54 \\
online-n2v(streamwalk)            &                0.64 &                0.80 &                0.52 &                0.53 &                0.47 &                0.39 &                0.56 &            13.95 \\
online-n2v(secondorder)           &                0.54 &                0.57 &                0.48 &                0.49 &                0.49 &                0.40 &                0.45 &            10.20 \\
\bottomrule
\end{tabular}

            }
        \end{sc}
    
\label{table:mean-gr-distortion-all}
\end{table*}

%% file: tables_new_fixed/calculation-time/mean-calculation-time.tex
\begin{table*}
    \centering
    \caption{Comparison of {\FrameworkName} and other methods in embeddings calculation time (in seconds). The presented values are the mean calculation time over 8 graph snapshots and 30 methods' retrains. We also report the mean ranks of all methods in this experiment. Methods marked in bold are the 3 best methods based on the mean rank. Underlined values show the lowest calcualtion time for each dataset individually.}
        \begin{sc}
            \scalebox{0.75}{
\begin{tabular}{lccccccc|c}
\toprule
&  bitcoin &  bitcoin &  fb &  fb &  enron & hypertext09 & radoslaw &  mean rank \\
& alpha & otc & forum & messages & employees & & email & \\
\midrule
n2v                               &               46.71 &              133.83 &               24.94 &               75.72 &                4.40 &                9.52 &               11.90 &            17.91 \\
line                              &               70.78 &               63.10 &              103.97 &              114.16 &              114.84 &              105.17 &               58.56 &            22.50 \\
ctdne                             &               16.25 &               61.41 &                3.98 &               77.44 &               16.82 &                4.39 &                2.65 &            15.86 \\
hope                              &                3.12 &                4.93 &                1.51 &                2.29 &                0.81 &                0.48 &                1.39 &             6.09 \\
dgi                               &                3.57 &               15.83 &                2.95 &                5.10 &                0.43 &                0.58 &                0.53 &             8.46 \\
le                                &                5.34 &               13.81 &                1.71 &                2.64 &                0.29 &                0.25 &                0.38 &             5.52 \\
lle                               &            $\times$ &            $\times$ &                4.22 &            $\times$ &                0.28 &                0.25 &                0.66 &             7.09 \\
\midrule
{\MethodNameBasicLC}(n2v)         &                7.87 &               22.26 &                8.44 &               27.62 &                2.35 &                4.79 &                4.65 &            13.29 \\
{\MethodNameBasicLC}(line)        &               42.34 &               41.66 &               80.53 &               82.18 &               84.43 &               83.29 &               43.50 &            19.79 \\
{\MethodNameBasicLC}(ctdne)       &                7.84 &               18.11 &                1.91 &               33.67 &                8.69 &                2.63 &                0.68 &            10.84 \\
$\textbf{{\MethodNameBasicLC}(hope)}$&                1.09 &  $\underline{1.52}$ &  $\underline{1.02}$ &  $\underline{1.35}$ &                0.34 &                0.25 &                0.46 &  $\textbf{2.71}$ \\
{\MethodNameBasicLC}(dgi)         &  $\underline{0.86}$ &                4.59 &                2.10 &                2.91 &                0.34 &                0.51 &                0.38 &             5.09 \\
$\textbf{{\MethodNameBasicLC}(le)}$&                1.24 &                2.01 &                1.45 &                1.68 &                0.25 &                0.19 &  $\underline{0.28}$ &  $\textbf{2.38}$ \\
$\textbf{{\MethodNameBasicLC}(lle)}$&            $\times$ &            $\times$ &                1.81 &            $\times$ &  $\underline{0.16}$ &  $\underline{0.11}$ &                0.33 &  $\textbf{2.34}$ \\
\midrule
{\MethodNameGeneralizedLC}(n2v)   &                8.61 &               24.04 &                9.90 &               30.69 &                3.71 &                5.31 &                6.40 &            14.50 \\
{\MethodNameGeneralizedLC}(line)  &               43.10 &               43.20 &               81.74 &               85.02 &               85.67 &               83.81 &               45.09 &            20.82 \\
{\MethodNameGeneralizedLC}(ctdne) &                8.59 &               19.88 &                3.33 &               36.37 &                9.96 &                3.15 &                2.71 &            13.61 \\
{\MethodNameGeneralizedLC}(hope)  &                1.78 &                3.50 &                2.24 &                4.14 &                1.42 &                0.75 &                1.91 &             8.21 \\
{\MethodNameGeneralizedLC}(dgi)   &                1.56 &                5.99 &                3.15 &                5.18 &                1.44 &                0.98 &                1.80 &             9.57 \\
{\MethodNameGeneralizedLC}(le)   &                1.94 &                3.42 &                2.48 &                3.96 &                1.31 &                0.65 &                1.67 &             7.70 \\
{\MethodNameGeneralizedLC}(lle)   &            $\times$ &            $\times$ &                2.80 &            $\times$ &                1.29 &                0.63 &                1.74 &             9.34 \\
\midrule
dynGraph2Vec(aernn)               &              144.71 &              341.14 &               25.54 &               84.10 &               15.89 &               13.35 &               15.47 &            18.54 \\
tNodeEmbed                        &               52.46 &              143.07 &               31.32 &               84.88 &               24.96 &               16.90 &               24.85 &            20.20 \\
online-n2v(streamwalk)            &               67.76 &              112.97 &               27.44 &               52.09 &              130.27 &               12.50 &              212.59 &            20.14 \\
online-n2v(secondorder)           &               20.78 &               72.98 &               70.51 &               74.15 &               76.83 &               20.92 &               86.89 &            19.70 \\
\bottomrule
\end{tabular}

            }
        \end{sc}
    
\label{table:mean-calculation-time-all}
\end{table*}

%% file: tables_new_fixed/mm/mean-mm.tex
\begin{table*}
    \centering
    \caption{Comparison of {\FrameworkName} and other methods in in max. memory utilization (in MB). The presented values are the mean max. memory utilization during embeddings calculation over 8 graph snapshots and 30 methods' retrains. We also report the mean ranks of all methods in this experiment. Methods marked in bold are the 3 best methods based on the mean rank. Underlined values show the lowest memory utilization for each dataset individually.}
        \begin{sc}
            \scalebox{0.75}{
\begin{tabular}{lccccccc|c}
\toprule
&  bitcoin &  bitcoin &  fb &  fb &  enron & hypertext09 & radoslaw & mean rank \\
& alpha & otc & forum & messages & employees & & email & \\
\midrule
n2v                               &                580.09 &                317.12 &                552.29 &                360.35 &                212.27 &                218.45 &                227.75 &             8.70 \\
line                              &                917.17 &                947.08 &               1291.09 &               1309.96 &               1286.30 &               1277.53 &                920.00 &            22.75 \\
ctdne                             &               1751.63 &               1262.22 &               1327.77 &               1113.66 &                344.43 &                320.13 &                361.23 &            19.18 \\
hope                              &                669.68 &                938.47 &                300.07 &                384.13 &                258.63 &                253.09 &                257.59 &            13.05 \\
dgi                               &                501.36 &                776.38 &                418.67 &                504.65 &                344.67 &                338.79 &                352.69 &            16.16 \\
le                                &                538.78 &                728.63 &                283.31 &                329.16 &                260.49 &                254.95 &                273.09 &            11.89 \\
lle                               &              $\times$ &              $\times$ &                259.21 &              $\times$ &                235.45 &                217.13 &                258.49 &             7.56 \\
\midrule
{\MethodNameBasicLC}(n2v)         &                233.96 &                344.62 &  $\underline{207.13}$ &                385.37 &  $\underline{206.60}$ &                241.49 &  $\underline{203.49}$ &             4.75 \\
{\MethodNameBasicLC}(line)        &                888.59 &                899.76 &               1275.47 &               1279.96 &               1271.36 &               1269.56 &                889.90 &            20.96 \\
{\MethodNameBasicLC}(ctdne)       &                436.55 &                849.80 &                338.78 &               1270.43 &                514.53 &                384.26 &                300.85 &            16.14 \\
$\textbf{{\MethodNameBasicLC}(hope)}$&                268.60 &                298.96 &                247.20 &                268.57 &                218.52 &                219.81 &                220.97 &  $\textbf{4.73}$ \\
{\MethodNameBasicLC}(dgi)         &                342.13 &                437.32 &                381.80 &                401.09 &                323.55 &                323.18 &                328.49 &            13.36 \\
$\textbf{{\MethodNameBasicLC}(le)}$&                247.31 &  $\underline{266.87}$ &                242.43 &  $\underline{247.96}$ &                220.95 &                219.35 &                220.60 &  $\textbf{3.43}$ \\
$\textbf{{\MethodNameBasicLC}(lle)}$&              $\times$ &              $\times$ &                228.67 &              $\times$ &                215.67 &  $\underline{213.14}$ &                216.45 &  $\textbf{2.62}$ \\
\midrule
{\MethodNameGeneralizedLC}(n2v)   &  $\underline{224.82}$ &                344.28 &                225.41 &                387.52 &                213.02 &                241.50 &                218.23 &             5.38 \\
{\MethodNameGeneralizedLC}(line)  &                888.82 &                901.10 &               1276.48 &               1281.39 &               1271.71 &               1269.49 &                889.21 &            21.18 \\
{\MethodNameGeneralizedLC}(ctdne) &                437.24 &                848.13 &                340.99 &               1272.02 &                514.99 &                385.14 &                300.77 &            16.43 \\
{\MethodNameGeneralizedLC}(hope)  &                265.25 &                300.65 &                260.65 &                290.92 &                232.45 &                228.96 &                241.79 &             6.79 \\
{\MethodNameGeneralizedLC}(dgi)   &                344.94 &                442.24 &                392.31 &                405.24 &                333.50 &                329.50 &                343.56 &            14.32 \\
{\MethodNameGeneralizedLC}(le)   &                252.32 &                267.62 &                261.33 &                265.87 &                233.62 &                224.25 &                238.77 &             5.61 \\
{\MethodNameGeneralizedLC}(lle)   &              $\times$ &              $\times$ &                255.57 &              $\times$ &                237.24 &                225.54 &                239.13 &             7.47 \\
\midrule
dynGraph2Vec(aernn)               &               1373.12 &               1934.64 &                899.86 &               1023.55 &                793.38 &                774.93 &                816.70 &            20.91 \\
tNodeEmbed                        &                643.19 &               1225.00 &                774.82 &                827.67 &                624.12 &                598.00 &                618.93 &            19.14 \\
online-n2v(streamwalk)            &                321.29 &                381.68 &                269.47 &                307.71 &                347.81 &                249.34 &                390.14 &            11.59 \\
online-n2v(secondorder)           &                260.86 &                284.38 &                258.63 &                265.96 &                253.60 &                246.26 &                259.93 &             7.61 \\
\bottomrule
\end{tabular}
            }
        \end{sc}
    
\label{table:mean-mm-all}
\end{table*}